\documentclass{article}

\PassOptionsToPackage{numbers, compress}{natbib}

\usepackage[final]{neurips_2022}

\usepackage[utf8]{inputenc} 
\usepackage[T1]{fontenc}    
\usepackage{hyperref}       
\usepackage{url}            
\usepackage{booktabs}       
\usepackage{amsfonts}       
\usepackage{nicefrac}       
\usepackage{microtype}      
\usepackage{amssymb}
\usepackage{pifont}
\newcommand{\xmark}{\ding{55}}%
\usepackage{wrapfig,lipsum,booktabs}
\usepackage{amsmath}
\usepackage[dvipsnames]{xcolor}
\usepackage{mathtools} 
\usepackage{multirow}
\usepackage{graphicx}  
\usepackage{color}
\usepackage{colortbl}
\usepackage{comment}
\usepackage{subfigure}
\usepackage{algorithm}
\usepackage{listings}

\makeatletter
\def\hlinewd#1{%
\noalign{\ifnum0=`}\fi\hrule \@height #1 \futurelet
\reserved@a\@xhline}

\definecolor{recolor}{rgb}{0,1,0}

\definecolor{srcolor}{rgb}{1,0,0}

\definecolor{shcolor}{rgb}{0,0,1}

\title{Integrative Feature and Cost Aggregation with Transformers for Dense Correspondence}

\newcommand*\samethanks[1][\value{footnote}]{}
\renewcommand\footnotemark{}

\author{
Sunghwan Hong$^{*}$\thanks{$^*$Equal contribution}\\
Korea University \\
\And
Seokju Cho$^{*}$\samethanks\\
Korea University \\
\And
Seungryong Kim$^{\dagger}$\thanks{$^\dagger$Corresponding author} \\
Korea University \\
\And
Stephen Lin\\
Microsoft Research Asia \\
}

\begin{document}

\maketitle


\begin{abstract}

We present a novel architecture for dense correspondence. The current state-of-the-art are Transformer-based approaches that focus on either feature descriptors or cost volume aggregation. However, they generally aggregate one or the other but not both, though joint aggregation would boost each other by providing information that one has but other lacks, \textit{i.e.,} structural or semantic information of an image, or pixel-wise matching similarity. In this work, we propose a novel Transformer-based network that interleaves both forms of aggregations in a way that exploits their complementary information. 
Specifically, we design a self-attention layer that leverages the descriptor to disambiguate the noisy cost volume and that also utilizes the cost volume to aggregate features in a manner that promotes accurate matching.
A subsequent cross-attention layer performs further aggregation conditioned on the descriptors of both images and aided by the aggregated outputs of earlier layers. We further boost the performance with hierarchical processing, in which coarser level aggregations guide those at finer levels. We evaluate the effectiveness of the proposed method on dense matching tasks and achieve state-of-the-art performance on all the major benchmarks. Extensive ablation studies are also provided to validate our design choices.  
\end{abstract}

\section{Introduction}
Finding visual correspondence between images is a central problem in computer vision, with numerous applications including simultaneous localization and mapping (SLAM)~\cite{bailey2006simultaneous}, augmented reality (AR)~\cite{peebles2021gan}, object tracking, structure from motion (SfM)~\cite{schonberger2016structure}, optical flow~\cite{fleet2006optical}, and image editing~\cite{barnes2009patchmatch,kim2018recurrent}. Given visually or semantically similar images, unlike sparse correspondence approaches~\cite{lowe2004distinctive,calonder2010brief,yi2016lift,detone2018superpoint,ono2018lf,dusmanu2019d2,revaud2019r2d2} that first detect a set of sparse points and extract corresponding descriptors to find matches across them, dense correspondence~\cite{truong2020glu,Hong_2021_ICCV,zheng2022dip,huang2022flowformer} aims at finding matches for all pixels. Dense correspondence approaches typically follow the classical matching pipeline~\cite{scharstein2002taxonomy,philbin2007object} of feature extraction, cost aggregation, and flow estimation. 

\begin{figure}[t]
\centering
\renewcommand{\thesubfigure}{}
\subfigure[(a) Feature aggregation]
{\includegraphics[width=0.35\textwidth]{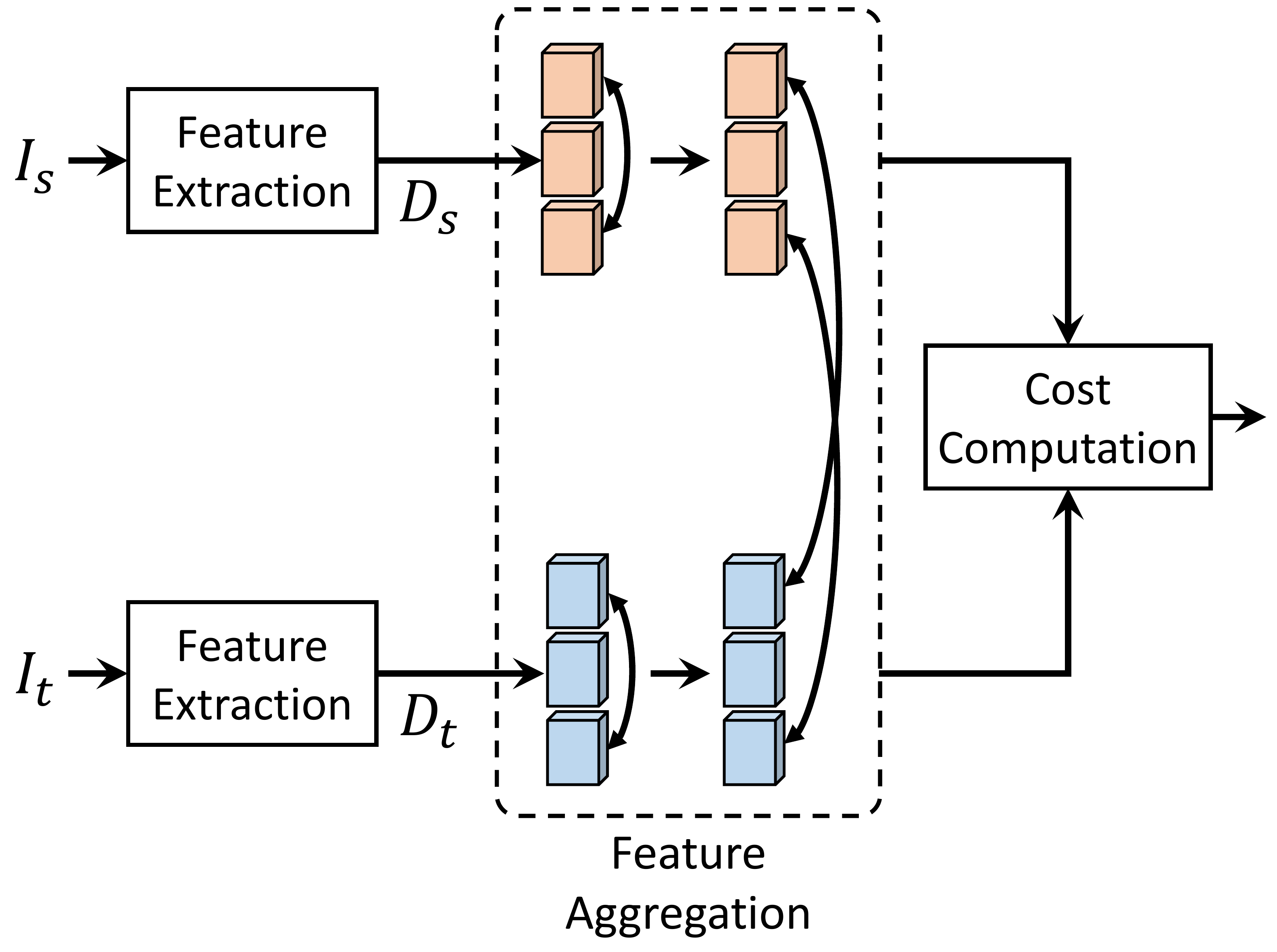}}\hfill
\subfigure[(b) Cost aggregation]
{\includegraphics[width=0.25\textwidth]{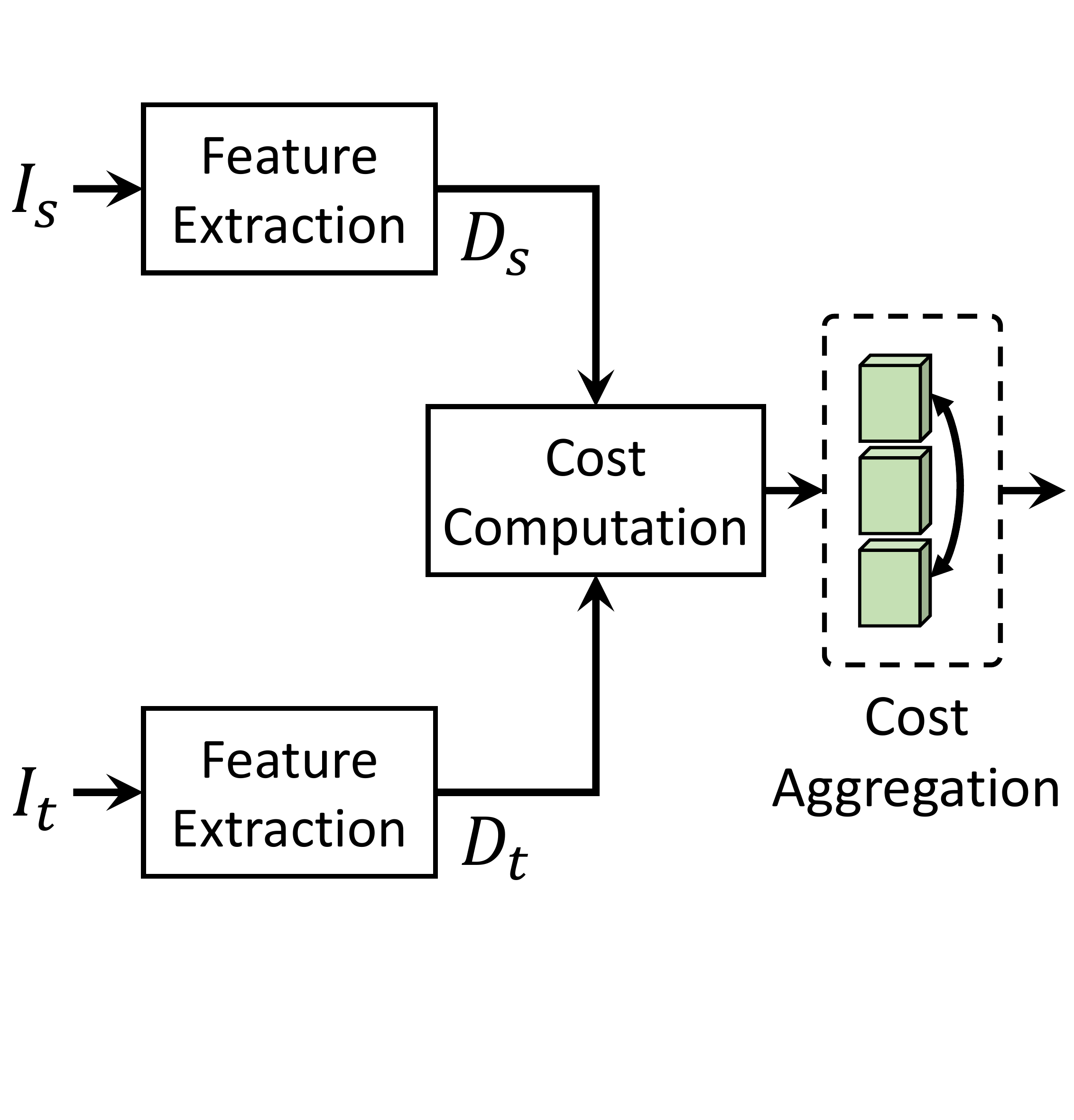}}\hfill
\subfigure[(c) Integrative aggregation (Ours)]
{\includegraphics[width=0.377\textwidth]{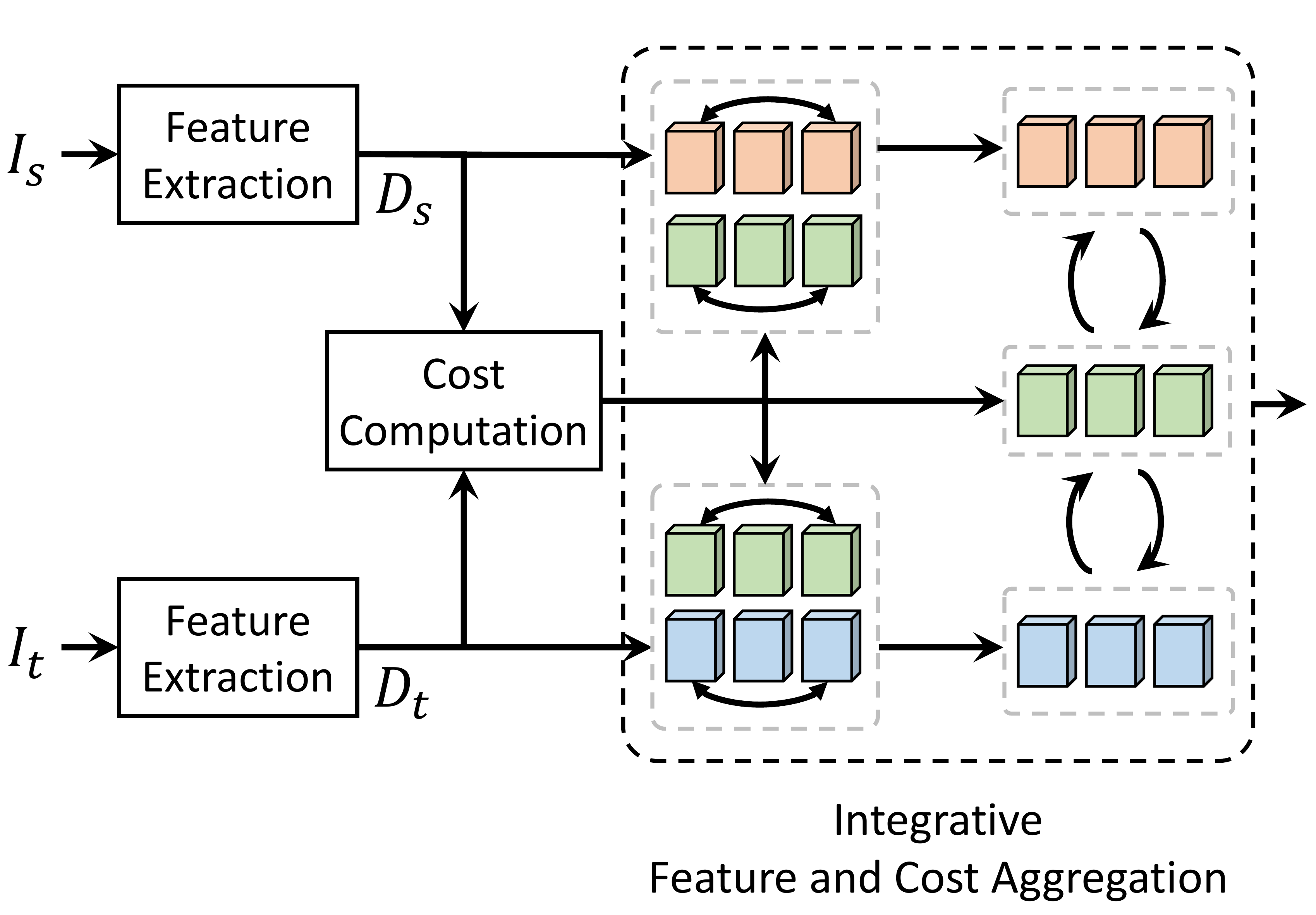}}\hfill\\
\vspace{-5pt}
\caption{\textbf{Intuition of the proposed method:} (a) feature aggregation methods~\cite{sun2021loftr,xu2021gmflow,jiang2021cotr} that aggregate feature descriptors, (b) cost aggregation methods~\cite{cho2021semantic,huang2022flowformer} that aggregate a cost volume, and (c) our integrative feature and cost aggregation method, which jointly performs both aggregations in a complementary manner.   }
\label{intuition}\vspace{-10pt}
\end{figure}

Much research has designed a means to address either the feature extraction or the cost aggregation, as shown in Fig.~\ref{intuition} (a) and (b). Feature aggregation aims to not only integrate self-similar features within an image and but also align similar features between the two images for matching, such as by using deep dense feature descriptors~\cite{lee2019sfnet,sarlin2020superglue,sun2021loftr,xu2021gmflow,jiang2021cotr}. The advantages of feature aggregation are particularly evident in attention and Transformer-based methods~\cite{vaswani2017attention,sarlin2020superglue,sun2021loftr,jiang2021cotr,xu2021gmflow} thanks to their attention layers with global receptive fields and adaptability to input tokens, which previous works with convolutions~\cite{Rocco18b,lee2019sfnet,jeon2020guided,Hong_2021_ICCV,min2021convolutional} lack. These methods, however, solely aggregate feature descriptors without consideration of cost aggregation. Numerous works~\cite{Rocco18b,huang2019dynamic,li2020correspondence,liu2020semantic,sarlin2020superglue,min2021convolutional,min2021convolutional++,cho2021semantic,sun2021loftr,hong2021cost,huang2022flowformer,cho2022cats++} on dense correspondence proposed methods for cost aggregation stage instead and demonstrate its importance. During cost aggregation, pair-wise interactions between pixels of the two images are considered by first computing a cost volume between descriptors and then suppressing noise to promote accurate correspondence estimation. Transformer-based methods~\cite{cho2021semantic,huang2022flowformer} are found to benefit significantly from cost aggregation, but they disregard aggregation of feature descriptors even though an improved cost volume constructed using less noisy features would ease the subsequent cost aggregation.

We argue that both feature aggregation and cost aggregation should ideally be performed in dense correspondence, as they serve different purposes and the benefits of each are well-established. Although there have been a few approaches~\cite{sarlin2020superglue,Hong_2021_ICCV} that attempt to aggregate both, no approaches utilized transformers due to its expensive computational complexity, and we believe that directly performing the two aggregations independently in a sequential manner only allows one aggregation to benefit the other but not vice versa, thus limiting the synergy between these two processes.

In this work, we present a method, which we call \textbf{I}ntegrative \textbf{F}eature and \textbf{C}ost \textbf{A}ggregation with \textbf{T}ransformers (\textbf{IFCAT}), that jointly aggregates feature descriptors and the cost volume in a manner that leverages their complementarity, as shown in Fig.~\ref{intuition} (c). 
This goal is accomplished in two steps, the first of which employs a self-attention layer to jointly aggregate the descriptors and cost volume. In this stage, the descriptors help to disambiguate the noisy cost volume similarly to cost volume filtering~\cite{hosni2012fast,sun2018pwc}, and the cost volume enhances feature aggregation by introducing matching similarities as a factor for aggregation. The cost volume explicitly represents the similarity of features in one image with respect to the features in the other, and accounting for it drives the features in each image to become more compatible with those of the other. In the subsequent step, we design a cross-attention layer that performs further aggregation aided by both the feature descriptors and the cost volume from earlier aggregations. By constructing better cross-attention maps with both the feature descriptors and the aggregated cost volume, the aggregated features of both images can be mutually improved more effectively. These self- and cross-attention layers are interleaved to facilitate convergence. We further boost performance through hierarchical processing that enhances this complementary aggregation by providing coarser outputs to guide finer-level aggregation.

We evaluate the proposed method on semantic and geometric matching tasks. In the experiments, we show that IFCAT outperforms prior works on all the major benchmarks, including SPair-71k~\cite{min2019spair}, PF-PASCAL~\cite{ham2017proposal}, PF-WILLOW~\cite{ham2016proposal} and HPatches~\cite{balntas2017hpatches}, by a significant margin, establishing a new state-of-the-art for all of them. We also conduct an extensive ablation study to validate our approach and the architectural design choices. The pre-trained weights and codes will be made available.

\section{Related Work}
\paragraph{Feature Extraction.}
Various methods have been proposed to extract feature descriptors for robust sparse matching. This process involves detecting interest points and extracting the descriptors of corresponding points. In traditional methods~\cite{liu2010sift,bay2006surf,dalal2005histograms,tola2009daisy}, the matching performance mostly relies on the quality of the feature detection and the description method, and outlier rejection across matched points is typically determined by RANSAC~\cite{fischler1981random}. These methods focus on the problem of identifying more meaningful feature points and extracting feature descriptors given an image. Despite their solid performance, they often struggle in cases of extreme appearance or viewpoint changes.

To overcome such issues, several methods~\cite{yi2016lift,detone2018superpoint,ono2018lf,dusmanu2019d2,revaud2019r2d2} extract dense deep features used to obtain descriptors tailored for matching. These works have demonstrated that the quality of feature descriptors contributes substantially to matching performance. In accordance with this, recent matching networks~\cite{detone2018superpoint,sarlin2020superglue,min2019hyperpixel,lee2019sfnet,Hong_2021_ICCV,min2020learning,jiang2021cotr,sun2021loftr,xu2021gmflow} proposed effective means for feature aggregation. 
Notable works include SuperGlue~\cite{sarlin2020superglue}, which employs graph self- and cross-attention to aggregate deep feature maps. SFNet~\cite{lee2019sfnet} and DMP~\cite{Hong_2021_ICCV} introduce an adaptation layer subsequent to feature extraction in order to learn feature maps well-suited to matching.

Recent state-of-the-art works utilize Transformer~\cite{vaswani2017attention} for feature aggregation. COTR~\cite{jiang2021cotr} uses transformers by taking input coordinates and feature maps to infer the correspondence of a given pixel coordinate, with self-attention computed for feature aggregation. LOFTR~\cite{sun2021loftr} also uses self- and cross-attention, but leaves the cost aggregation to a handcrafted method, \textit{i.e.,} optimal transport~\cite{sinkhorn1967diagonal}. Very recently, GMFlow~\cite{xu2021gmflow} also utilized a transformer for feature aggregation in optical flow estimation. Despite its state-of-the-art performance, its disregard of cost aggregation may lead to sub-optimal solutions, a problem we address in this paper. 
\vspace{-5pt}

\paragraph{Cost Aggregation.}
In the correspondence literature, recent works carefully design their architecture to effectively aggregate a cost volume. Some works~\cite{dosovitskiy2015flownet,sun2018pwc,hui2018liteflownet,melekhov2019dgc,truong2020glu,Hong_2021_ICCV,jeon2020guided,truong2021learning} use 2D convolutions to establish correspondence while aggregating the cost volume with learnable kernels that have a local receptive field. Although 2D convolutions are used for flow estimation, they in fact also aggregate costs during the process, making them suitable for both the cost aggregation and flow estimation stages. Some works~\cite{min2019hyperpixel,min2020learning,liu2020semantic} utilize handcrafted methods including RHM~\cite{cho2015unsupervised} and the OT solver~\cite{sinkhorn1967diagonal}. These works have inherent limitations as their use of handcrafted techniques do not take advantage of learning and are susceptible to severe deformations. NC-Net~\cite{Rocco18b} proposes to use 4D convolutions for cost aggregation in order to identify sets of consistent matches by exploring neighborhood consensus. Inspired by this, numerous works~\cite{li2020correspondence,yang2019volumetric,huang2019dynamic,rocco2020efficient,min2021convolutional,min2021convolutional++} either adopted or extended 4D convolutions. For example, DCCNet~\cite{huang2019dynamic} used it for cost embedding; Sparse NC-Net~\cite{li2020correspondence} designed adaptive 4D convolutions, and~\cite{yang2019volumetric,rocco2020efficient,min2021convolutional++} proposed efficient versions. However, they are all limited in the sense that they inherit the limitations of CNN-based architectures, for which the receptive fields are local.

Recently, CATs~\cite{cho2021semantic} proposed to use Transformer~\cite{vaswani2017attention} as a means for cost aggregation, and an extension~\cite{cho2022cats++} combined convolutions with Transformer for enhanced cost aggregation. Although they benefit from the global receptive field of self-attention operations, they disregard feature aggregation even though the cost volume they use is constructed from feature maps. FlowFormer~\cite{huang2022flowformer} takes an approach that utilizes Transformer, but it is designed specifically for the optical flow task and does not aggregate features. By disregarding feature aggregation, these methods may limit their performance due to the resultant noise in the cost volume which then hampers cost aggregation.
\vspace{-5pt}


\section{Preliminaries: Self- and Cross-Attention}

Self- and cross-attention are the core elements of Transformer~\cite{vaswani2017attention} for their ability to globally model relationships and interactions among input tokens and their adaptability to input tokens. As a general description, given a sequence of tokens as an input, Transformer~\cite{vaswani2017attention} first linearly projects tokens to obtain query, key and value embeddings. These are then fed into a scaled dot product attention layer, followed by layer normalization (LN)~\cite{ba2016layer} and a feed-forward network or MLP, to produce an output with the same shape as the input. Each token is attended to by all the other tokens. This attention process can be formulated as:
\begin{equation}
\begin{split}
        Q = \mathcal{P}_Q(X),\quad
        K = \mathcal{P}_K(X), \quad
        V = \mathcal{P}_V(X),
\end{split}
\end{equation}
where $\mathcal{P}_Q$, $\mathcal{P}_K$ and $\mathcal{P}_V$ denote query, key and value projections, respectively, and $X$ denotes a token with a positional embedding. The obtained query, key and value embeddings then pass through an attention layer: 
\begin{equation}
    \mathrm{Attention}(X) = \mathrm{softmax}(\frac{QK^T}{\sqrt{d_k}})V,
    \label{eq:2}
\end{equation}
where $d_k$ is the dimension of key embeddings. Note that the $\mathrm{Attention}(\cdot)$ function can be defined in various ways~\cite{wang2020linformer,liu2021swin,katharopoulos2020transformers,lu2021soft,wu2021fastformer}. A key factor that distinguishes self- and cross-attention is the input to the key and value projections. Given a pair of input tokens, e.g., $X_s$ and $X_t$, the input to the key and value projections when performing self-attention with $X_s$ is the same input, $X_s$, but for cross-attention across $X_s$ and $X_t$, the inputs to the key and value projection are $X_t$.


\section{Methodology}\label{sec:3}

\subsection{Problem Formulation}
Let us denote a pair of visually or semantically similar images, i.e., the source and target, as $I_{s}$ and $I_{t}$, the feature descriptors extracted from $I_{s}$ and $I_{t}$ as $D_{s}$ and $D_{t}$, respectively, and the cost volume computed between the feature maps as $C$. Given $I_s$ and $I_t$, we aim to establish a correspondence field ${F}(i)$ that is defined at all pixels $i$ and warps $I_s$ towards $I_t$. 

Recent learning-based networks~\cite{rocco2017convolutional,Rocco18b,rocco2020efficient,cho2021semantic} accomplish dense correspondence by extracting features from deep CNNs~\cite{he2016deep} or Transformers~\cite{dosovitskiy2020image} for $D_s$ and $D_t$. The extracted features subsequently undergo $l$-2 normalization. A cost volume that consists of all pair-wise feature similarities $C \in \mathbb{R}^{h \times w \times h \times w}$ with height $h$ and width $w$ is then computed and stored: $C(i,j)=D_{s}(i)\cdot {D}_{t}(j)$,
where $i$ and $j$ index the source and target features, respectively. 
To improve the matching performance, existing state-of-the-art methods perform either feature aggregation~\cite{sun2021loftr,xu2021gmflow} or cost aggregation~\cite{cho2021semantic,huang2022flowformer} with Transformer~\cite{vaswani2017attention} such that $\{{D}_{s}',D_{t}'\}=\mathcal{T}(D_s,D_t)$ or $C' = \mathcal{T}(C)$, where $\mathcal{T}(\cdot)$ denotes Transformer. Then, ${F}(i)$ is determined from $C(i,j)$ considering all $j$.

\subsection{Motivation and Overview}

We argue that solely focusing on either feature or cost aggregation may lead to sub-optimal solutions. While feature aggregation~\cite{sun2021loftr,xu2021gmflow} is a process of aligning similar features based on the rich structural or semantic information present in dense feature maps, cost aggregation~\cite{huang2022flowformer,cho2021semantic} is a process of suppressing noises based on matching similarities. The information the two aggregations consider are different, but they can enhance matching by improving each other's aggregation with the help of information that the other has. 

In this work, we aim to jointly learn feature and cost aggregation modules by establishing a complementary relationship between them. To this end, we first concatenate a cost volume and feature descriptors, which we call a feature cost volume, and feed it to the self-attention layer. Within the self-attention layer, both feature and cost aggregation are performed, where the feature descriptors and the cost volume benefit from one another during the aggregation. Subsequently, we leverage the aggregated features and cost volume for cross-attention, which performs further aggregation aided by the aggregated inputs to the cross-attention layer. These self- and cross-attention layers are interleaved to facilitate convergence. Finally, we formulate our architecture in a coarse-to-fine manner, where the outputs of coarser attention blocks are used to guide the aggregation of finer-level blocks. In the following, we will explain each module in detail.
\begin{figure*}
    \centering
    \includegraphics[width=1\linewidth]{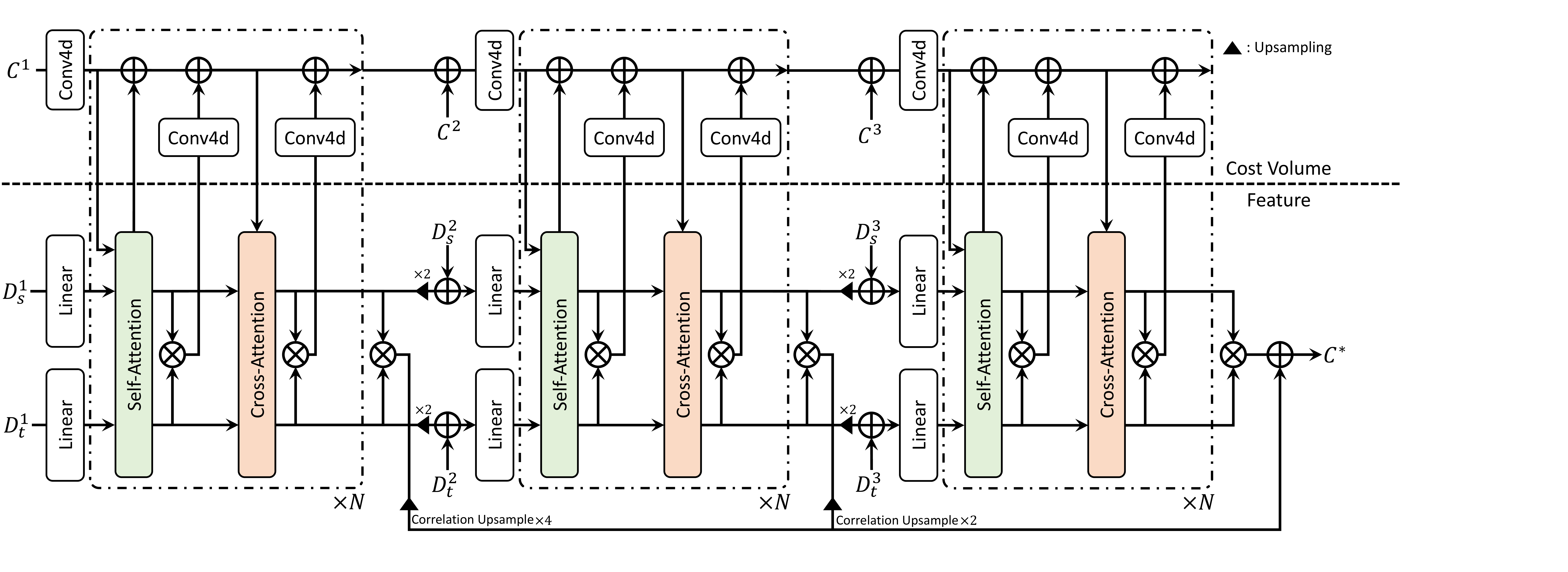}\hfill\\
    \caption{\textbf{Overall architecture of the proposed method.} Given feature maps $D_s$ and $D_t$, and cost volume $C$ as inputs, our method employs self- and cross-attention specifically designed to conduct joint feature aggregation and cost aggregation in a coarse-to-fine manner.}
    \label{fig:overall}\vspace{-10pt}
\end{figure*}

\subsection{Integrative Feature and Cost Aggregation}
\paragraph{Self-Attention for Integrative Feature and Cost Aggregation.}
Subsequent to feature extraction and cost computation, we feed both feature descriptors $D_s,$ $D_t$ and cost volume $C$ into our proposed self-attention layer. We first embed these inputs with linear projection for channel reduction prior to self-attention, and positional embeddings~\cite{vaswani2017attention} are added after the query and key projections, as shown in Fig.~\ref{fig:overall}. As done in~\cite{Rocco18b}, to ensure input order invariance, we consider the bidirectional nature of cost volumes and feed a pair $\{D_s,C\}$ and $\{D_t,C^\mathrm{T}\}$ into the proposed self-attention layer independently, where $C^\mathrm{T}(i,j) = C(j,i)$, and add the outputs. 

\begin{figure*}[t]
\centering
\renewcommand{\thesubfigure}{}
\subfigure[(a) Proposed self-attention]
{\includegraphics[width=0.5756091954022988\textwidth]{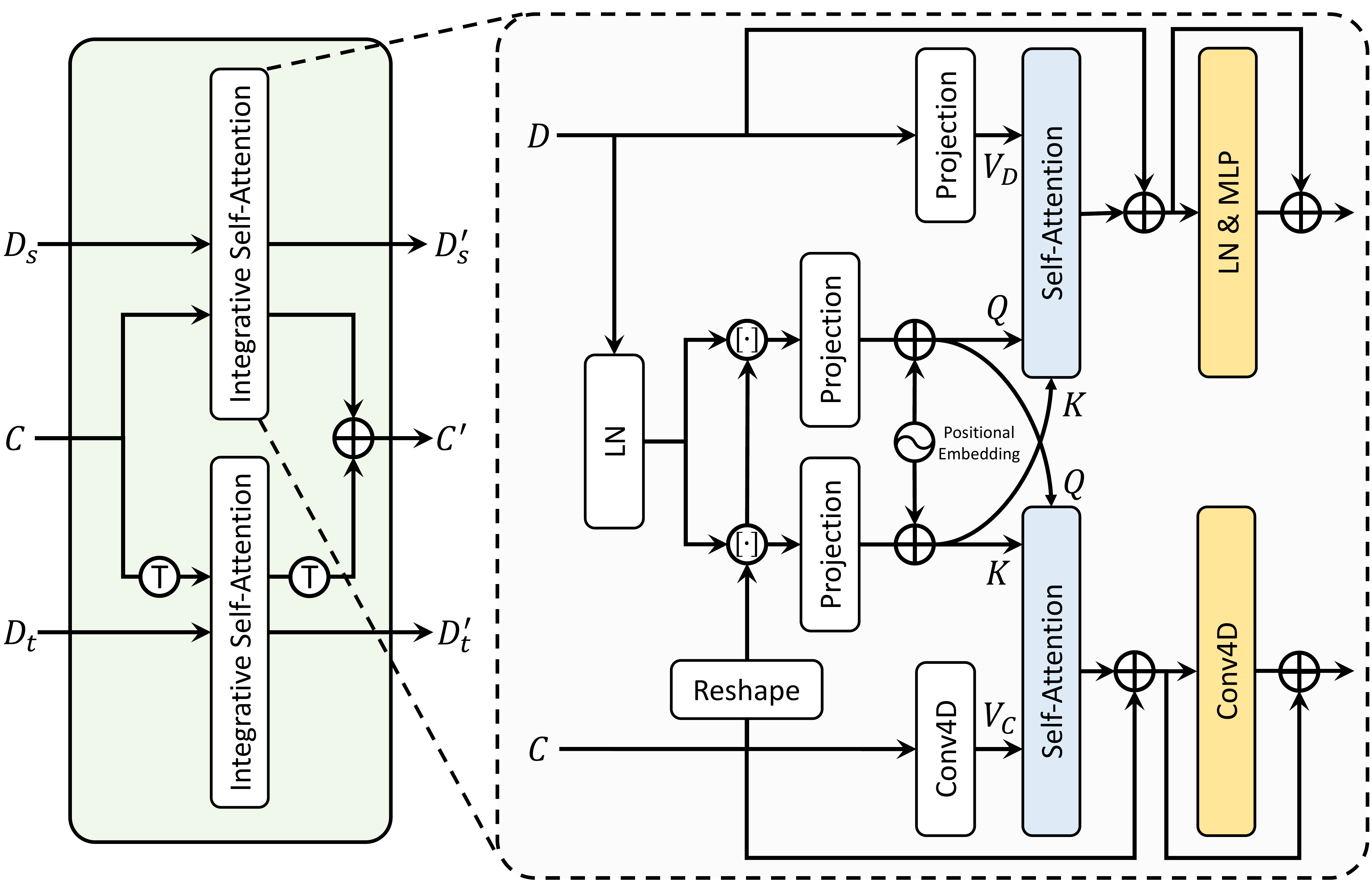}}\hfill
\subfigure[(b) Proposed cross-attention]
{\includegraphics[width=0.4043908045977011\textwidth]{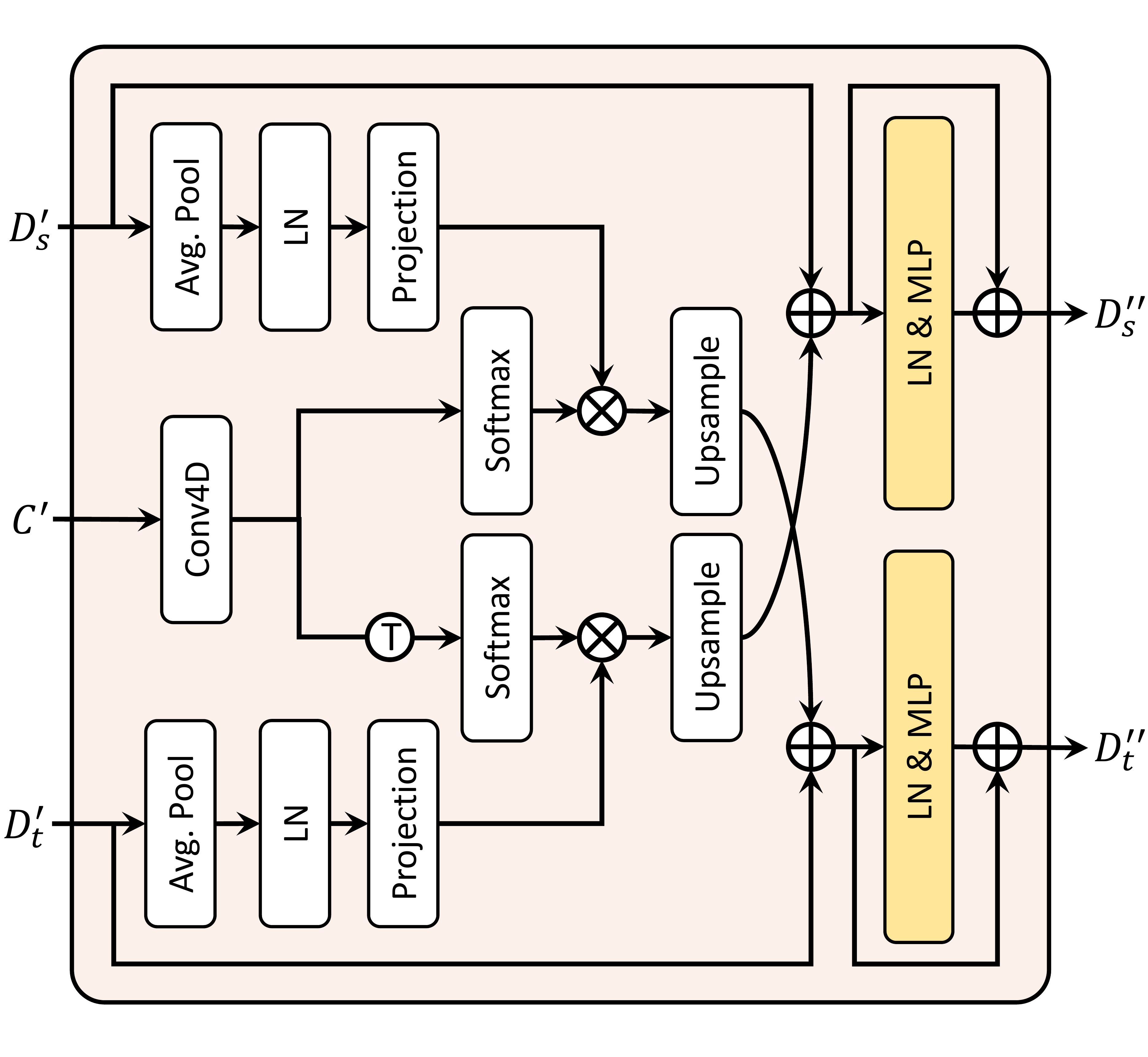}}\hfill
\\
\vspace{-5pt}
\caption{\textbf{Illustration of the proposed self- and cross-attention:} (a) self-attention layer that performs joint feature aggregation and cost aggregation, and (b) cross-attention layer with matching distribution for enhanced feature aggregation. }
\label{fig:attention}\vspace{-10pt}
\end{figure*}

Specifically, as shown in Fig.~\ref{fig:attention}, within the self-attention layer, we first obtain a feature cost volume $[D,C]$ by concatenating $D$ and $C$, where $[\,\cdot\,]$ denotes concatenation. To compute self-attention, we need to define the query, key and value embeddings. Unlike other works~\cite{sun2021loftr,cho2021semantic} that solely aggregate either the feature descriptors or cost volume, we jointly aggregate both. To this end, we exploit the feature cost volume in computing self-attention and define two independent value embeddings, specifically one for feature projection and the other for cost volume projection. Note that we do not use the feature cost volume for value embeddings to ensure disentangled aggregation that is targeted for one and not the other. Formally, we define the query, key and values as:
\begin{equation}
    \begin{split}
Q = \mathcal{P}_{Q}([D , C]), \quad K = \mathcal{P}_{K}([D , C]), \quad  V_{D} = \mathcal{P}_{V_D}(D), \quad V_{C} = \mathcal{P}_{V_C}(C),
    \end{split}
\label{eq:cost_volume}
\end{equation}    
where $V_D$ and $V_C$ denote the value embeddings of feature descriptors and the cost volume, respectively. After computing an attention map by applying softmax over the query and key dot product, we use it to aggregate feature $D$ and cost volume $C$ with $V_D$ and $V_C$ using Eq.~\ref{eq:2} as follows:
\begin{equation}
    \mathrm{Attention}_\mathrm{self-D}(C,D) = \mathrm{softmax}(\frac{QK^T}{\sqrt{d_k}})V_D, 
    \quad 
    \mathrm{Attention}_\mathrm{self-C}(C,D) = \mathrm{softmax}(\frac{QK^T}{\sqrt{d_k}})V_C.
\end{equation}
Note that any type of attention computation can be utilized,~\textit{i.e.,} additive~\cite{bahdanau2014neural} or dot product~\cite{vaswani2017attention}, while in practice we use the linear kernel dot product with the associative property of matrix products~\cite{katharopoulos2020transformers}. The outputs of this self-attention are denoted as $D'_s$, $D'_t$, and $C'$.

In this way, we benefit from two perspectives. From the cost aggregation point of view, the feature map of the feature cost volume can disambiguate the noisy cost volume as proven in the stereo matching literature~\cite{yoon2006adaptive,hosni2012fast,he2011global},~\textit{i.e.,} 
cost volume filtering, which aids the cost aggregation process. From the feature aggregation point of view, the cost volume explicitly represents the similarity of features in one image with respect to the features in the other, and accounting for it drives the features in each image to become more compatible with those of the other. 
\vspace{-5pt}

\paragraph{Cross-Attention with Matching Distribution.}
In the proposed cross-attention layer, the aggregated feature and cost volume are explicitly used for further aggregation, and we condition both feature descriptors on both input images via this layer. By exploiting the outputs of the self-attention layer, the cross-attention layer performs cross-attention between feature descriptors for further feature aggregation using the improved feature descriptors $D'_s$, $D'_t$ and enhanced cost volume $C'$ from earlier aggregations. 

Concretely, as shown in Fig.~\ref{fig:attention}, we first treat the input cost volume as a cross-attention map, since applying a softmax function over the cost volume is tantamount to obtaining an attention map. 
In this way, we could perform more enhanced aggregation as the input cost map to the cross-attention layer is guided by residual connections that contain information from improved features and an enhanced cost volume as further explained in the next paragraph. Note that features $D'_s$ and $D'_t$ undergo average pooling followed by layer normalization~\cite{ba2016layer} and a linear projection layer in order to adjust the spatial resolution to $C'$, as will be further explained in Section~\ref{coarse-to-fine}.
Formally, we first define a cross-attention map and value for attention score as $QK^T = C'$ and $V_{D'} = \mathcal{P}_{V_D}(D')$, respectively. The attention process for cross-attention is then defined as follows:
\begin{equation}
    \mathrm{Attention}_\mathrm{cross}(C',D') = \mathrm{softmax}(\frac{C'}{\sqrt{d_k}})V_{D'}.
    \label{2}
\end{equation}
The outputs of this cross-attention are denoted as $D''_s$, $D''_t$, and $C''$.
\vspace{-5pt}

\begin{figure*}[t]
\centering
\renewcommand{\thesubfigure}{}
\subfigure[(a) Source]
{\includegraphics[width=0.124\textwidth]{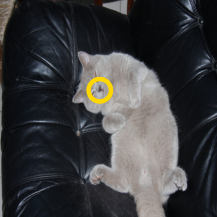}}\hfill
\subfigure[(b) Target]
{\includegraphics[width=0.124\textwidth]{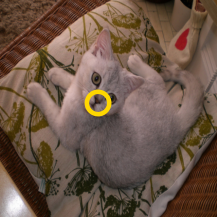}}\hfill
\subfigure[(c) $C^1$ ]
{\includegraphics[width=0.124\textwidth]{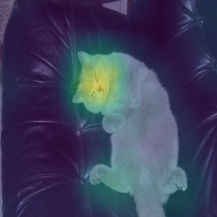}}\hfill
\subfigure[(d) $C^2$]
{\includegraphics[width=0.124\textwidth]{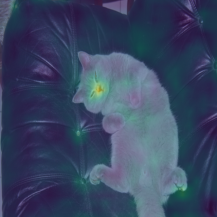}}\hfill
\subfigure[(e) $C^3$]
{\includegraphics[width=0.124\textwidth]{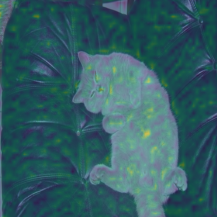}}\hfill
\subfigure[(f) $D'_s$ $\cdot$ $D'_t$]
{\includegraphics[width=0.124\textwidth]{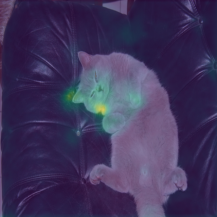}}\hfill
\subfigure[(g) $C'$]
{\includegraphics[width=0.124\textwidth]{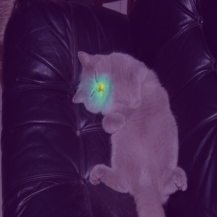}}\hfill
\subfigure[(h) $D''_s$ $\cdot$ $D''_t$]
{\includegraphics[width=0.124\textwidth]{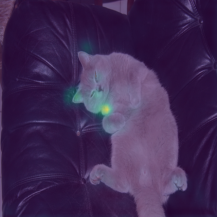}}\hfill
\vspace{-5pt}\\
\caption{\textbf{Visualization of attention maps:} (a) source image, (b) target image, (c)-(e) raw correlations, (f) a cost volume constructed with features aggregated with self-attention, (g) a cost volume aggregated with self-attention, and (h) a cost volume constructed with features aggregated with cross-attention.}
\label{fig:multi-cost}\vspace{-10pt}
\end{figure*}


\paragraph{Enhanced Aggregation with Improved Features and Cost Volume.}
Within the proposed attention block, it is shown in Fig.~\ref{fig:overall} that the outputs of the self- and cross-attention layers, which include aggregated feature maps and a cost volume, are connected to the next layer or added to the input cost volume. Specifically, the feature maps are used in two ways: for cost volume construction and as inputs to aggregation in subsequent layers. For cost volume construction, the aggregated features are processed according to Eq.~\ref{eq:cost_volume} and the result is added to the cost volume input of the next layer by a residual connection. This is repeated for both self-attention and cross-attention layers across $N$ attention blocks, where the aggregated pairs of one block are fed as input to the following attention block. 
Similar to the feature maps, the cost volume is progressively improved from aggregation and better features as it passes through the attention blocks. 
Through these mechanisms of the proposed joint aggregation, the feature maps and cost volume help each other in self-attention and also facilitate further aggregation by providing an enhanced cross-attention map and improved features. 

\subsection{Coarse-to-Fine Formulation }\label{coarse-to-fine}
To improve the robustness of fine-scale estimates, we extend our architecture to a coarse-to-fine approach through pyramidal processing, as done in~\cite{jeon2018parn,melekhov2019dgc,truong2020glu,Hong_2021_ICCV}. Specifically, we use feature maps of the last index at each pyramid level, specifically feature maps from $\mathrm{conv3_x}$ to $\mathrm{conv5_x}$ when ResNet~\cite{he2016deep} is used.
We first use a coarse pair of refined feature maps and aggregated cost volume, and similar to~\cite{zhao2021multi} that learns complementary correspondence by adding the cost volume of the previous scale, we progressively learn complementary descriptors and correspondences through adding the previous-level outputs to those of the next level.

The overall architecture is shown in Fig.~\ref{fig:overall}. A straightforward solution would be to upsample all the outputs of the cross-attention layer, following a pyramidal structure. However, the increasing computational and memory burden with respect to the cost volume resolution makes this infeasible. To alleviate such issues, we fix the resolution of the cost volume to the coarsest resolution and utilize 4D convolutions to 
downsample the spatial resolution when finer cost volumes are added to the coarsest cost volume. 
Finally, we sum up the output cost volumes after each attention block computed using the enhanced feature maps across all levels for the final output of the network. Formally, given the outputs of the attention block at each level, ${D}''^{,l}_s, {D}''^{,l}_t$ and ${C}''^{,l}$, where $l$ denotes the $l$-th level, we upsample the aggregated features using bilinear interpolation, and add them to the raw feature descriptors extracted from $I_s$ and $I_t$ defined at the next level: ${D}^{{l+1}}_s = {D}^{{l+1}}_s + \mathrm{up}({D}''^{,l}_s)$, where ${D}^{{l+1}}_t$ is defined similarly. In addition, ${C}^{l+1} = {C}^{l+1} + \mathrm{Conv4d}({C}''^{,l})$. The visualizations of cost volumes are shown in Fig.~\ref{fig:multi-cost}.

Finally, given the features ${D}''_s$ and ${D}''_t$ at each level, we compute the correlation map between ${D}''_s$ and ${D}''_t$, and the sum of cost volumes across all levels are added up to obtain the final output ${C}^*$ that is used to estimate the final flow field, as shown in the bottom of Fig.~\ref{fig:overall}.

\subsection{Training Objective}
To train the networks, we first compute the correlation map between ${D}''_s$ and ${D}''_t$ at each level and then transform it into a dense flow field $F_\mathrm{pred}$ using the soft-argmax operator~\cite{lee2019sfnet}. Then, we compare the predicted dense flow field with the ground-truth flow field $F_\mathrm{GT}$. Specifically, we use average end-point error (AEPE), computed by averaging the Euclidean distance between the ground-truth and estimated flow, for the objective function. We then sum up the AEPE loss across all levels. We thus formulate the objective function as $\mathcal{L}= \|F_\mathrm{GT}-F_\mathrm{pred}\|_{2}$. Flow fields may instead be obtained from ${C}''$, but we empirically find that slightly better results are obtained when feature maps are used.

\begin{figure*}[t]
\centering
\renewcommand{\thesubfigure}{}
\subfigure[Source $I_s$ \ \ \ \ \ \ \ \ \ \ \ \ \ \ \ \ \ \ \ \ \ \ \ \ \ \ \ \ Target $I_t$]
{\includegraphics[width=0.499\textwidth]{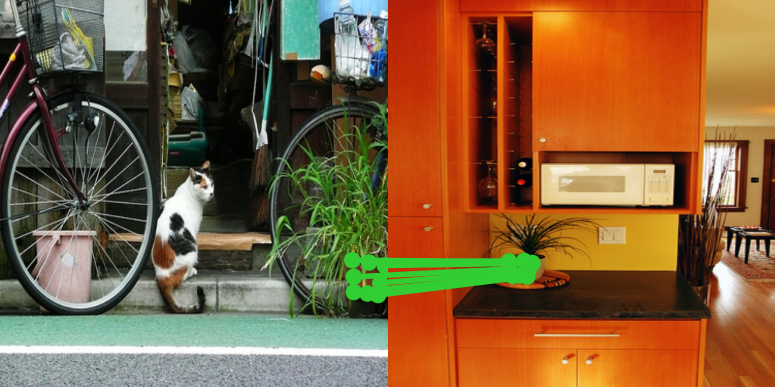}}\hfill
\subfigure[Source $I_s$ \ \ \ \ \ \ \ \ \ \ \ \ \ \ \ \ \ \ \ \ \ \ \ \ \ \ \ \ Target $I_t$]
{\includegraphics[width=0.499\textwidth]{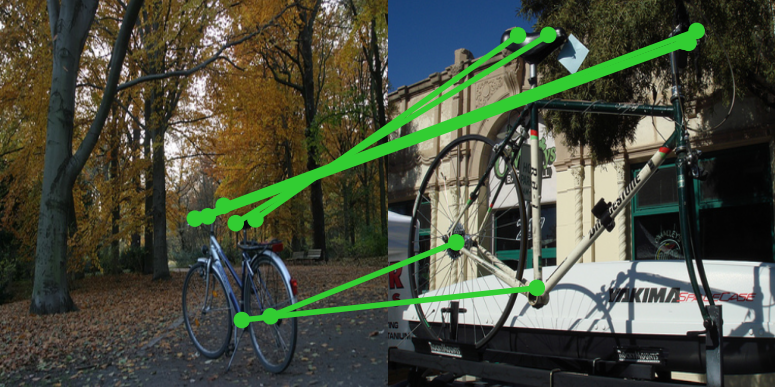}}\hfill\\
\vspace{-5pt}
\subfigure[Source $I_s$]
{\includegraphics[width=0.124\textwidth]{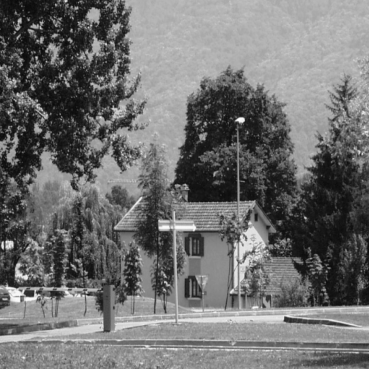}}\hfill
\subfigure[Target $I_t$]
{\includegraphics[width=0.124\textwidth]{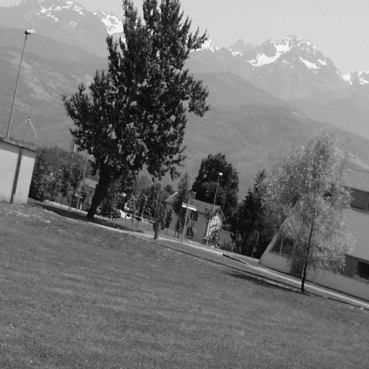}}\hfill
\subfigure[GT]
{\includegraphics[width=0.124\textwidth]{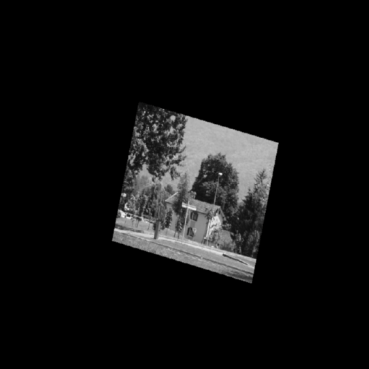}}\hfill
\subfigure[Prediction]
{\includegraphics[width=0.124\textwidth]{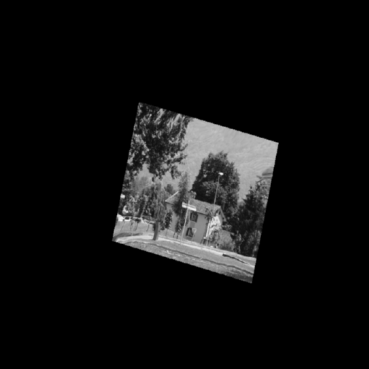}}\hfill
\subfigure[Source $I_s$]
{\includegraphics[width=0.124\textwidth]{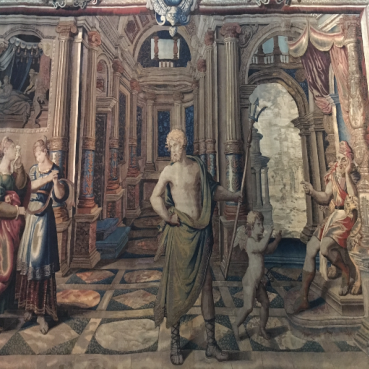}}\hfill
\subfigure[Target $I_t$]
{\includegraphics[width=0.124\textwidth]{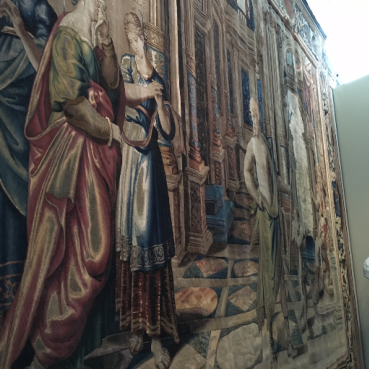}}\hfill
\subfigure[GT]
{\includegraphics[width=0.124\textwidth]{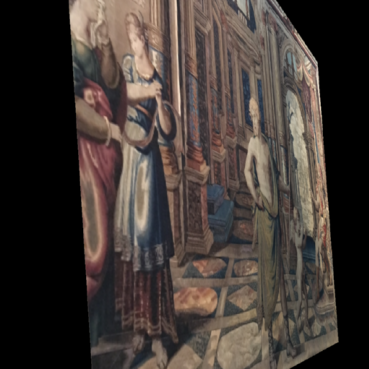}}\hfill
\subfigure[Prediction]
{\includegraphics[width=0.124\textwidth]{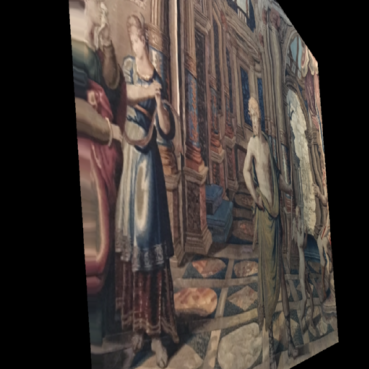}}\hfill\\
\vspace{-5pt}
\caption{\textbf{Qualitative results on SPair-71k~\cite{min2019spair} (top) and HPatches~\cite{balntas2017hpatches} (bottom).}}
\label{fig:qualitative}\vspace{-10pt}
\end{figure*}

\section{Experiments}
\subsection{Implementation Details}\label{sec:4.1}
For the backbone network, we use VGG-16~\cite{simonyan2014very} for dense alignment and ResNet-101~\cite{he2016deep} for dense semantic correspondence, which are both pretrained on ImageNet~\cite{deng2009imagenet}. We use N$ = (2,2,2)$ for efficiency in training. We use data augmentation used in~\cite{cho2021semantic} for dense semantic correspondence, while no augmentation is used for dense alignment. Our networks are trained with the input images resized to 512$\times$512. We implemented our network using PyTorch~\cite{paszke2017automatic}, and the AdamW~\cite{loshchilov2017decoupled} optimizer is employed with an initial learning rate of 1e$-$4 for the IFCAT layers, which we gradually decrease using step learning rate decay. We train our networks for 50 epochs on SPair-71k~\cite{min2019spair} and DPED-CityScape-ADE~\cite{truong2020glu}, and 300 epochs for PF-PASCAL~\cite{ham2017proposal}. More details can be found in the supplementary material. The code and pretrained weights will be made publicly available.

\begin{table*}[!t]
    \begin{center}
    \caption{\textbf{Quantitative evaluation on standard semantic correspondence benchmarks~\cite{min2019spair,ham2016proposal,ham2017proposal}.}  Higher PCK is better. The best results are in bold, and the second best results are underlined. \textit{Reso.: Resolution, F.A.: Feature Aggregation, C.A.: Cost Aggregation.} 
    }\label{tab:main_table}\vspace{+5pt}
    \scalebox{0.67}{
    \begin{tabular}{l|c|c|c|c|ccc|cc|cc}
            \hlinewd{0.8pt}
             \multirow{3}{*}{Methods}&\multirow{3}{*}{Reso.} &\multirow{3}{*}{F.A.} & \multirow{3}{*}{C.A.}& SPair-71k~\cite{min2019spair} & \multicolumn{3}{c|}{PF-PASCAL~\cite{ham2017proposal}} & \multicolumn{4}{c}{PF-WILLOW~\cite{ham2016proposal}} \\
          
              &&& &PCK @ $\alpha_{\text{bbox}}$ & \multicolumn{3}{c|}{PCK @ $\alpha_{\text{img}}$}  & \multicolumn{2}{c}{PCK @ $\alpha_{\text{bbox}}$} & \multicolumn{2}{c}{PCK @ $\alpha_{\text{bbox-kp}}$} \\ 
        
              & & & &0.1 & 0.05 & 0.1 & 0.15 & 0.05 & 0.1 & 0.05 &0.1 \\ 
             \midrule\midrule
           
             CNNGeo~\cite{rocco2017convolutional} &Ori&-&2D Conv. &20.6 &41.0 &69.5 &80.4 &- &- &36.9&69.2\\
             A2Net~\cite{seo2018attentive} &-&-&2D Conv. &22.3 &42.8 &70.8 &83.3 &- &- &36.3&68.8\\
              WeakAlign~\cite{rocco2018end} &Ori&-&2D Conv. &20.9 &49.0 &74.8 &84.0 &- &- &37.0&70.2\\
              RTNs~\cite{kim2018recurrent} &- & -& 2D Conv. &25.7 &55.2 &75.9 &85.2 &- &- &41.3&71.9\\ 
              SFNet~\cite{lee2019sfnet}&288/Ori & 2D Conv. & - &- &53.6 &81.9 &90.6 &- &- &46.3&74.0\\

             NC-Net~\cite{Rocco18b} &240/ori/400  &-&4D Conv. &20.1 &54.3 &78.9 &86.0 &- &- &33.8&67.0\\
             
             DCC-Net~\cite{huang2019dynamic}&240/ori/-  &- &4D Conv. &- &55.6 &82.3 &90.5 &- &- &43.6&73.8\\
             
             HPF~\cite{min2019hyperpixel}&Max 300 &- &RHM&28.2 &60.1 &84.8 &92.7 &- &- &45.9&74.4\\
             GSF~\cite{jeon2020guided}&- & -&2D Conv. &36.1&65.6 &87.8 &95.9 &- &- &\underline{49.1}&\textbf{78.7}\\
             ANC-Net~\cite{li2020correspondence} &-  &-&4D Conv. &- &- &86.1 &- &- &- &-&-\\
             DHPF~\cite{min2020learning}&240 &-&RHM &37.3 &75.7 &90.7 &95.0 &49.5 &77.6 &- &71.0\\
             SCOT~\cite{liu2020semantic}&Max 300& -&OT-RHM &35.6 &63.1 &85.4 &92.7 &- &- &{47.8}&{76.0}\\
             CHM~\cite{min2021convolutional} &240 & -& 6D Conv. &46.3 &80.1 &91.6 &94.9 &{52.7} &{79.4} &-&69.6\\
             PMNC~\cite{lee2021patchmatch} &400 &- & 4D Conv. & 50.4& \underline{82.4}& 90.6& -& -&- &-&-\\

             
             CATs~\cite{cho2021semantic} &256  &-&Trans.&49.9 &75.4 &\underline{92.6} &\underline{96.4} &50.3 &79.2 & 40.7&69.0\\
                  MMNet-{FCN}~\cite{zhao2021multi} &224$\times$320  &Conv.+Trans.& - & 50.4&81.1 & 91.6& 95.9&- &- &-&-\\
               PMD~\cite{li2021probabilistic} &- & Attention&-&37.4&-&90.7&-&-&-&-&75.6\\
              PWarpC-NC-Net~\cite{truong2022probabilistic} &Ori  &-&4D Conv.&{52.0}&79.2&92.1&95.6&-&-&48.0&\underline{76.2}\\
              
             VAT~\cite{hong2022cost} &512  &-&Conv.+Trans.&\underline{55.5} &78.2 &{92.3} &{96.2} &\underline{52.8} &\textbf{81.6} & -&-\\
      
          \midrule
             \textbf{IFCAT} (Ours) &Ori  &Trans. &Trans.&\textbf{64.4} &\textbf{88.0} &\textbf{94.8} &\textbf{97.9} &\textbf{58.6} &\underline{81.2} &\textbf{50.4}&74.2\\
             
            \hlinewd{0.8pt}
    \end{tabular}
    }
    \end{center}\vspace{-10pt}
\end{table*}

\subsection{Dense Semantic Correspondence}

In this section, we evaluate the effectiveness of the proposed method for dense correspondence. For a fair comparison, following~\cite{kim2018recurrent,huang2019dynamic,min2019hyperpixel,min2020learning,cho2021semantic}, when evaluating on SPair-71k~\cite{min2019spair} we train the proposed method on SPair-71k~\cite{min2019spair}, and when evaluating on PF-PASCAL~\cite{ham2017proposal} and PF-WILLOW~\cite{ham2016proposal} we train on PF-PASCAL~\cite{ham2017proposal}.

SPair-71k~\cite{min2019spair} consists of 70,958 image pairs with extreme and diverse viewpoints, scale variations, and rich annotations for each image pair. PF-PASCAL~\cite{ham2016proposal} contains 1,351 image pairs over 20 object categories with keypoint annotations, and PF-WILLOW~\cite{ham2016proposal} is composed of 900 image pairs from 4 categories. We use percentage of correct keypoints (PCK) for the evaluation metric, for which higher values are better. Concretely, given predicted keypoint $k_\mathrm{pred}$ and ground-truth keypoint $k_\mathrm{GT}$, we count the number of predicted keypoints that satisfy the following condition: $d( k_\mathrm{pred},k_\mathrm{GT}) \leq \alpha \cdot \mathrm{max}(H,W)$, where $d(\,\cdot\,)$ denotes Euclidean distance; $\alpha$ denotes a threshold value; $H$ and $W$ denote height and width of the object bounding box or the entire image. Note that as confirmed in~\cite{truong2022probabilistic,cho2022cats++}, using different ground-truth resolutions for evaluation leads to different results, so we set our evaluation on the original resolution.

The results are summarized in Table~\ref{tab:main_table}. As shown, IFCAT clearly sets a new state-of-the-art for the three dense semantic correspondence benchmarks. This demonstrates the effectiveness of joint aggregation by outperforming methods that focus on either feature or cost aggregation. 
 \vspace{-5pt}

\begin{table}[]
    \centering
     \caption{\textbf{Quantitative evaluation on HPatches~\cite{balntas2017hpatches}.} We evaluate on both HPatches-240 and HPatches original. Lower AEPE is better. We divide methods into two groups: multiple feed-forward and single feed-forward.}
    \scalebox{0.56}{
    \begin{tabular}{l|c|c|ccccc|c|c|ccccc|c|c}
    \toprule
    \multirow{3}{*}{Methods}&\multirow{3}{*}{F.A.}&\multirow{3}{*}{C.A.}&\multicolumn{7}{c|}{HPatches (240 $\times$ 240)} &\multicolumn{7}{c}{HPatches}\\\cline{4-17}
    &&&\multicolumn{6}{c|}{AEPE $\downarrow$}&\multirow{2}{*}{PCK}&\multicolumn{6}{c|}{AEPE $\downarrow$}&\multirow{2}{*}{PCK}\\\cline{4-9}\cline{11-16}
    &&&I&II&III&IV&V&Avg.& &I&II&III&IV&V&Avg.&\\\midrule\midrule
    COTR~\cite{jiang2021cotr} &Trans. &\xmark &-&-&-&-&-&-&-&-&-&-&-&-&\textbf{7.75}&\textbf{91.10}
			\\
			RANSAC-Flow~\cite{shen2020ransac} &2D Conv. &\xmark  &\textbf{0.51}&\underline{2.36}&\underline{2.91}&\textbf{4.41}&\bf{5.12}&\underline{3.06} &-&-&-&-&-&-&-&- \\
			RANSAC-DMP~\cite{Hong_2021_ICCV} &2D Conv. &2D Conv. &\underline{0.53}&\textbf{2.21}&\textbf{2.76}&\underline{4.62}&\underline{5.14}&\textbf{3.05}&\textbf{96.28}&\textbf{4.32}&\textbf{11.21}&\textbf{22.80}&\textbf{31.34}&\textbf{33.64}&\underline{20.65}&\textbf{75.35}
			\\\hline
         CNNGeo~\cite{rocco2017convolutional} &\xmark  &2D Conv.&9.59&18.55&21.15&27.83&35.19&22.46&-&-&-&-&-&-&-&- \\
         DGC-Net~\cite{melekhov2019dgc}  &\xmark  &2D Conv. &1.74&5.88&9.07&12.14&16.50&9.07&50.01&5.71&20.48&34.15&43.94&62.01&33.26&58.06 \\
			GLU-Net~\cite{truong2020glu} &\xmark &2D Conv. &\textbf{0.59}&\underline{4.05}&\underline{7.64}&\underline{9.82}&\underline{14.89}&\underline{7.40}&\underline{83.47}&\underline{1.55}&12.66&27.54&32.04&52.47&25.05&{78.54} \\
			GOCOR-GLU-Net~\cite{truong2020gocor} &\xmark&Hand-crafted &-&-&-&-&-&-&-&\textbf{1.29}&\textbf{10.07}&\underline{23.86}&\underline{27.17}&\underline{38.41}&\underline{20.16}&\textbf{81.43} \\

			DMP~\cite{Hong_2021_ICCV} &2D Conv. &2D Conv. &1.21&5.12&12.31&13.68&16.12&9.69&79.21&3.21&15.54&32.54&38.62&63.43&30.64&63.21
			\\\midrule\midrule
				
		\textbf{IFCAT} (Ours)&Trans. &Trans. &\underline{0.65}&\textbf{3.33}&\textbf{5.41}&\textbf{6.91}&\textbf{10.09}&\textbf{5.27}&\textbf{90.90}&1.90&\underline{10.72}&\textbf{18.95}&\textbf{24.36}&\textbf{31.40}&\textbf{17.59}&\underline{80.41}\\\bottomrule
    \end{tabular}}
    \label{tab:hpatches}\vspace{-10pt}
\end{table}

\subsection{Dense Alignment}

We further evaluate our method on the dense alignment dataset HPatches~\cite{balntas2017hpatches}. For each scene, there is a source image and five target images from different viewpoints, along with corresponding ground-truth flows. The resolutions of HPatches range from 450 $\times$ 600 to 1,613 $\times$ 1,210. As in~\cite{melekhov2019dgc}, we also evaluate on downsampled HPatches~\cite{balntas2017hpatches}, where the images are resized to a low resolution (240 $\times$ 240). For the evaluation metric, we use the average end-point error (AEPE), computed by averaging the Euclidean distance between the ground-truth and estimated flow, and percentage of correct keypoints (PCK), computed as the ratio of estimated keypoints within a threshold of the ground truth to the total number of keypoints. 
When evaluating on HPatches~\cite{balntas2017hpatches}, following~\cite{truong2020glu,truong2020gocor} we train our networks on the DPED-CityScape-ADE~\cite{truong2020glu} dataset.

Table~\ref{tab:hpatches} summarizes the quantitative results. As shown, IFCAT outperforms existing works by a large margin, clearly achieving state-of-the-art performance. Note that a fair comparison to COTR~\cite{jiang2021cotr} is not feasible because its use of the zoom-in technique with a multiple inference strategy dramatically boosts its performance as a trade-off to speed. It is shown that RANSAC-Flow~\cite{shen2020ransac} and RANSAC-DMP~\cite{Hong_2021_ICCV} exceeds the performance of our method, but they adopt two-stage inference where the first stage aims to find the homographic transformation between an image pair, which give them an advantage on the HPatches dataset. On its own, IFCAT is demonstrated to be effective, highlighting the importance of integrative aggregation.   

\subsection{Ablation Study}\label{sec:4.5}

\paragraph{Aggregation Strategy.}
In this ablation study, we compare the performance of different aggregation strategies. Table~\ref{tab:aggre} summarizes the results. There are seven components we evaluate. From (\textbf{I}) to (\textbf{II}), we report the results for sole aggregation on either feature or cost volume. We investigate the effectiveness of conditioning features on both images and performing both aggregations in (\textbf{III} and (\textbf{IV}).  Lastly, we progressively add the proposed components in (\textbf{V}) to (\textbf{VII}) to demonstrate their significance. For a fair comparison, we trained all strategies with a single level except for (\textbf{VII}).

\begin{wraptable}{r}{7.5cm}
\vspace{-5mm}
\caption{\textbf{Aggregation strategies for IFCAT.}}
\label{tab:aggre}\vspace{+5pt}
\centering
\resizebox{\linewidth}{!}{%
\begin{tabular}{ll|cc}
        \hlinewd{0.8pt}
        &\multirow{2}{*}{Components} &SPair-71k &HPatches \\
        &&$\alpha_{\text{bbox}}$ = 0.1 $\uparrow$ &AEPE $\downarrow$\\
        \midrule
        \textbf{(I)} &Feature self-att.  & 36.1&84.08\\
        \textbf{(II)} & Cost self-att.  & 28.7&54.80\\
        \textbf{(III)} &  Feature self-att. + cross-att. &38.5 & 81.72\\
        \textbf{(IV)} &  Sequential (\textbf{III}) + cost self-att. &56.5&49.00\\
        \textbf{(V)} &   Integrative self-att. &54.7 &34.85\\
        \textbf{(VI)}&Integrative self- and cross-att. &58.5&24.41\\
        \textbf{(VII)}& (\textbf{VI}) + hierarchical processing &64.4&17.59\\
        \hlinewd{0.8pt}
\end{tabular}%
}
\end{wraptable}
As shown, solely aggregating either the feature or cost volume yields limited performance, where (\textbf{II}) can be seen as a simplified version of CATs~\cite{cho2021semantic}. However, we observe that combining self- and cross-attention, which is highly similar to LoFTR~\cite{sun2021loftr}, for feature aggregation helps to boost the performance by conditioning features on both images. This highlights the importance of providing information from the other image, which implies that providing a cost volume that explicitly represents similarity information with respect to each image would help to establish a more accurate correspondence field. Interestingly, (\textbf{IV}) shows that performing feature and cost aggregation yields a large performance boost, demonstrating that cost aggregation benefits from powerful feature representations tailored to matching. From (\textbf{V}) to (\textbf{VII}), each component contributes appreciably to the improvement of performance, clearly showing the effectiveness of our proposed components. This confirms that, as the result of (\textbf{III}) implies, leveraging the complementarity of features and cost volume is of prime importance.  
\vspace{-5pt}

\paragraph{Depth of Attention Block.}
As shown in Fig.~\ref{fig:overall}, we can stack the attention block at each level to increase model capacity and allow further aggregation. In this ablation study, we show the effects of varying the hyperparameter $N$. We additionally show comparisons of memory consumption, run time, and number of learnable parameters to indicate their efficiency.

\begin{wraptable}{r}{8.5cm}
\vspace{-5mm}
\caption{\textbf{Ablations on varying depth of attention block.}}
\label{tab:depth}\vspace{+5pt}
\centering
\resizebox{\linewidth}{!}{%
\begin{tabular}{l|cc|ccc}
        \hlinewd{0.8pt}
        \multirow{2}{*}{\# of N} &SPair-71k &HPatches &Memory &Runtime &\# of param.\\
        &$\alpha_{\text{bbox}}$ = 0.1 $\uparrow$  &AEPE $\downarrow$ &[MB]&[ms]&[M]\\
        \midrule
        (1,0,0) &55.4&35.02&335.09&35.09&0.58\\
        (2,0,0) &58.5&24.41&354.57&47.23&1.04\\ 
        (1,1,1) &63.7&20.49&845.06&91.96&0.83\\ 
        (2,2,2) &64.4&17.59&874.25&138.05&1.55\\ 
        (3,3,3) &64.7&17.43&903.40&188.43&2.26\\ 
        \hlinewd{0.8pt}
\end{tabular}%
}
\end{wraptable}
The results are summarized in Table~\ref{tab:depth}. We consistently observe that as the attention block depth increases, the performance is boosted, demonstrating that the earlier aggregations help the subsequent aggregations.  We generally observe increasing memory consumption, run-time and number of parameters as $N$ increases, which are trade-offs to improve performance. Note that during training, the number of parameters has a direct influence on memory consumption, which is a limitation of increasing the attention block depth.


\section{Conclusion}

In this paper, we proposed a novel transformer-based architecture, called Integrative Feature and Cost Aggregation with Transformer (IFCAT), that interleaves feature refinement and cost aggregation by establishing a complementary relationship. Our design based on self- and cross-attention is tailored for matching and joint enhancement of feature descriptors and the cost volume. This method is formulated in a coarse-to-fine manner, yielding an appreciable performance boost. We have shown that our method surpasses all other existing works on several benchmarks for semantic and geometric matching, establishing new state-of-the-art performance. We also conducted an extensive ablation study to validate our choices.

\medskip
{\small
\bibliographystyle{plain}
\bibliography{egbib}
}

\newpage
\begin{center}
	\textbf{\Large Appendix}
\end{center}

\appendix
\renewcommand{\thefigure}{\arabic{figure}}
\renewcommand{\theHfigure}{A\arabic{figure}}
\renewcommand{\thetable}{\arabic{table}}
\renewcommand{\theHtable}{A\arabic{table}}
\setcounter{figure}{0}
\setcounter{table}{0}

In this document, we provide more implementation details and psuedo-code of IFCAT,  and more results on SPair-71k~\cite{min2019spair}, PF-PASCAL~\cite{ham2017proposal}, PF-WILLOW~\cite{ham2016proposal} and HPatches~\cite{balntas2017hpatches}.

\section*{More Details}

\paragraph{Training Details.}
For training, we employ the same augmentation strategy introduced in~\cite{cho2021semantic}. To implement 4D convolutions, we use separable 4D convolutions for efficient computation, which is introduced in VCN~\cite{yang2019volumetric}. All separable 4D convolutions are followed by ReLU activation and Layer Normalization~\cite{ba2016layer}. We set the weight decay to 0.05 and the learning rate to 1e$^{-4}$ for IFCAT and to 1e$^{-6}$ for the backbone, which is halved at epochs 30 and 40.

\paragraph{Psuedo-code.}
We present Pytorch-like Psuedo-code of the proposed method in Alg.~\ref{1}.

\begin{figure}[!t]
    \centering
    \renewcommand{\thesubfigure}{}
    \subfigure[]
	{\includegraphics[width=0.247\linewidth]{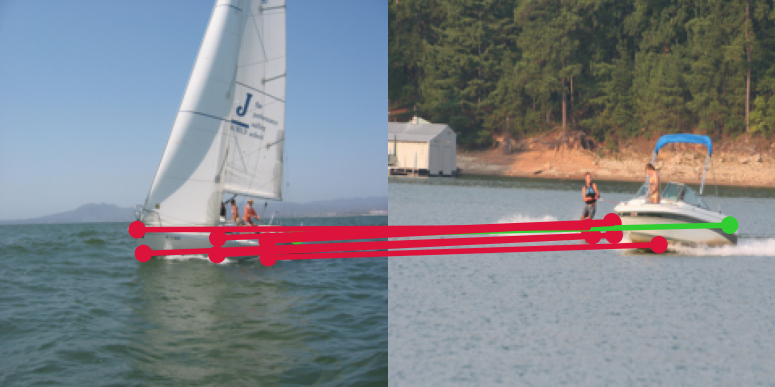}}\hfill
    \subfigure[]
	{\includegraphics[width=0.247\linewidth]{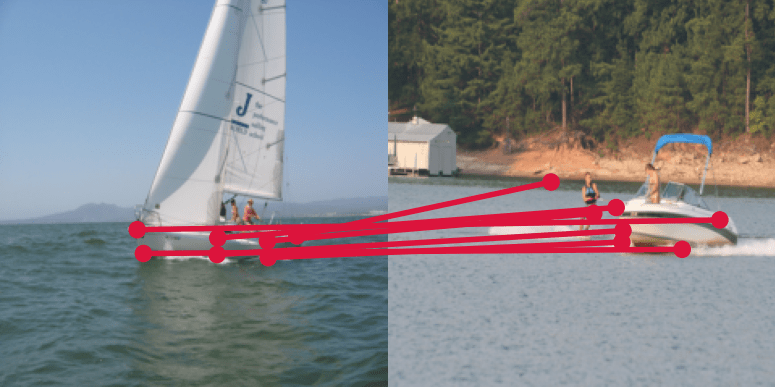}}\hfill
    \subfigure[]
	{\includegraphics[width=0.247\linewidth]{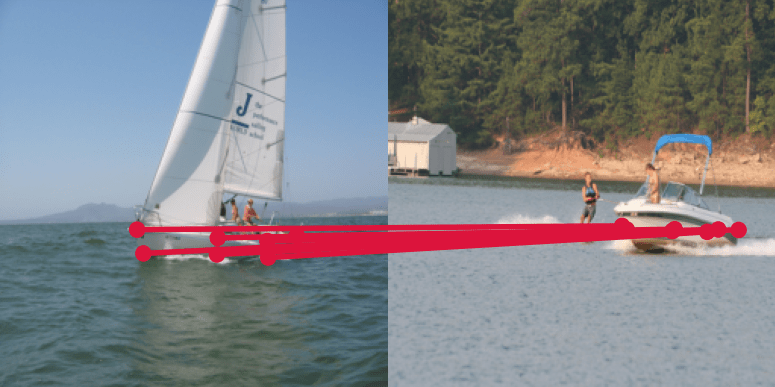}}\hfill
    \subfigure[]
	{\includegraphics[width=0.247\linewidth]{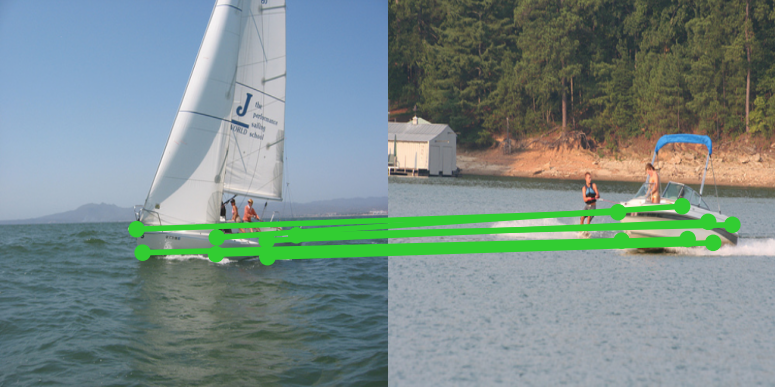}}\hfill\\
	\vspace{-20.5pt}
    \subfigure[]
	{\includegraphics[width=0.247\linewidth]{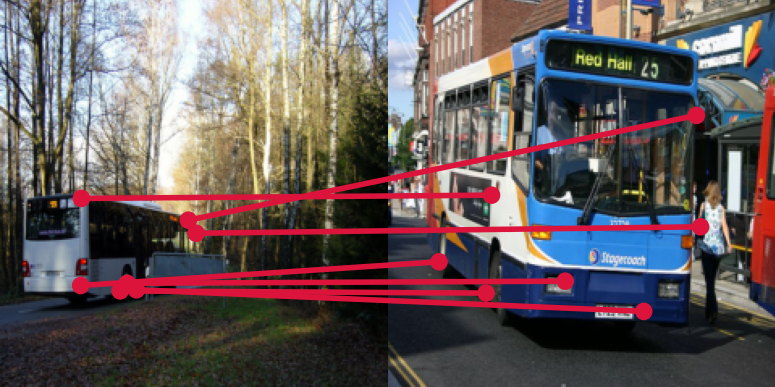}}\hfill
    \subfigure[]
	{\includegraphics[width=0.247\linewidth]{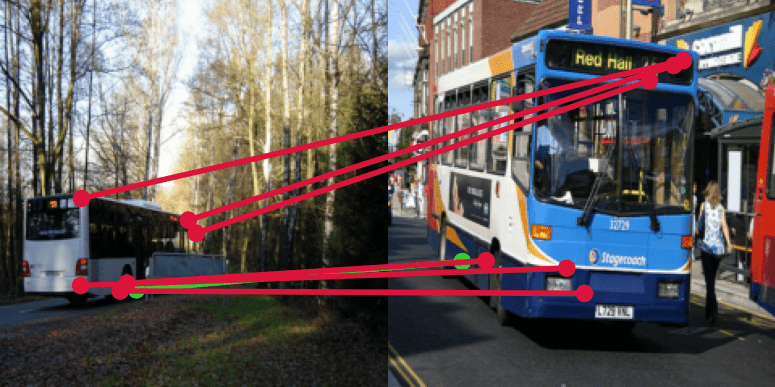}}\hfill
    \subfigure[]
	{\includegraphics[width=0.247\linewidth]{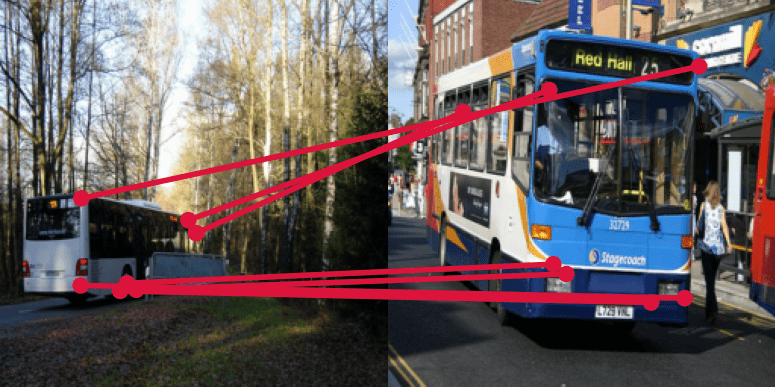}}\hfill
    \subfigure[]
	{\includegraphics[width=0.247\linewidth]{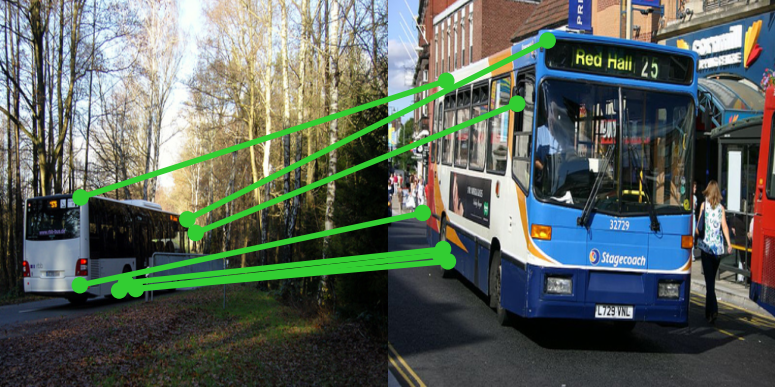}}\hfill\\
	\vspace{-20.5pt}
    \subfigure[]
	{\includegraphics[width=0.247\linewidth]{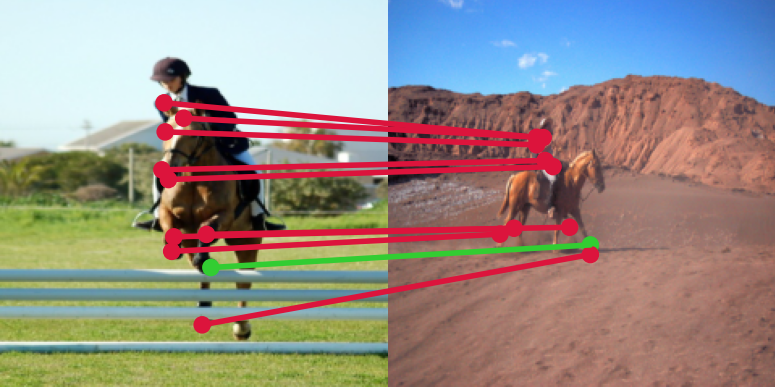}}\hfill
    \subfigure[]
	{\includegraphics[width=0.247\linewidth]{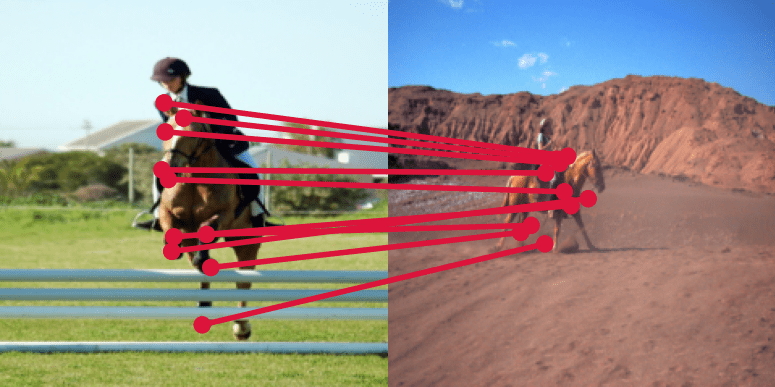}}\hfill
    \subfigure[]
	{\includegraphics[width=0.247\linewidth]{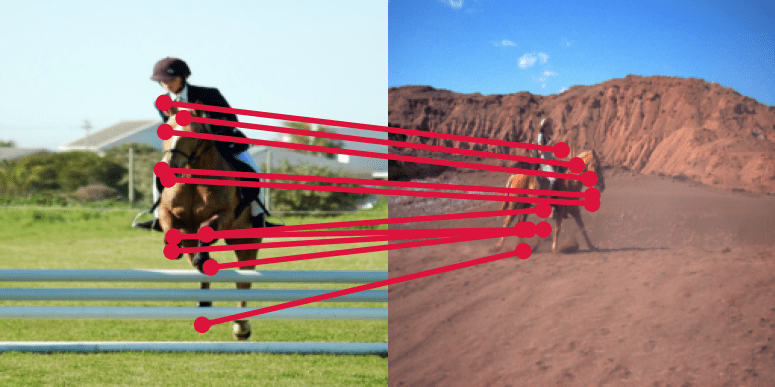}}\hfill
    \subfigure[]
	{\includegraphics[width=0.247\linewidth]{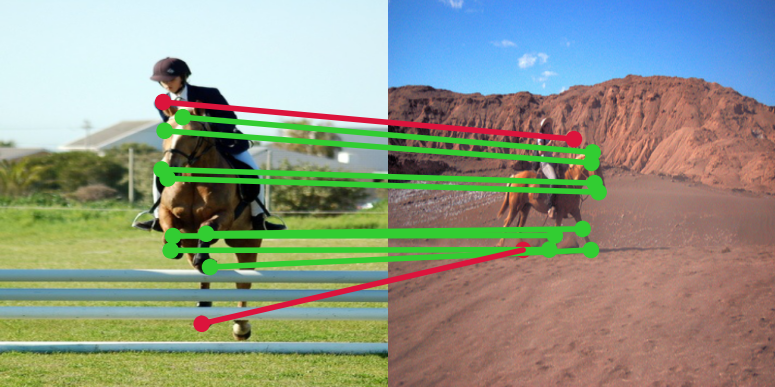}}\hfill\\
	\vspace{-20.5pt}
    \subfigure[]
	{\includegraphics[width=0.247\linewidth]{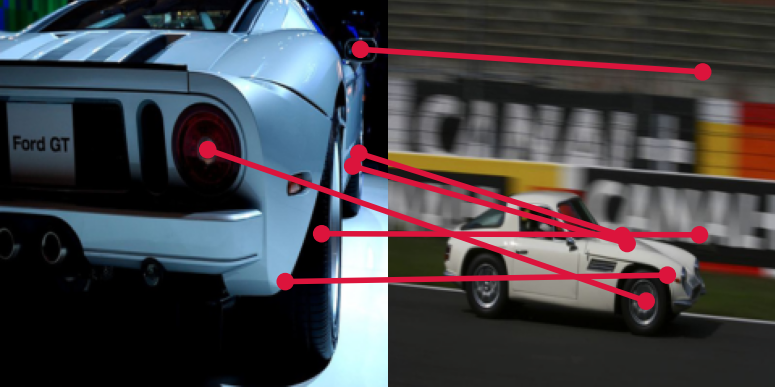}}\hfill
    \subfigure[]
	{\includegraphics[width=0.247\linewidth]{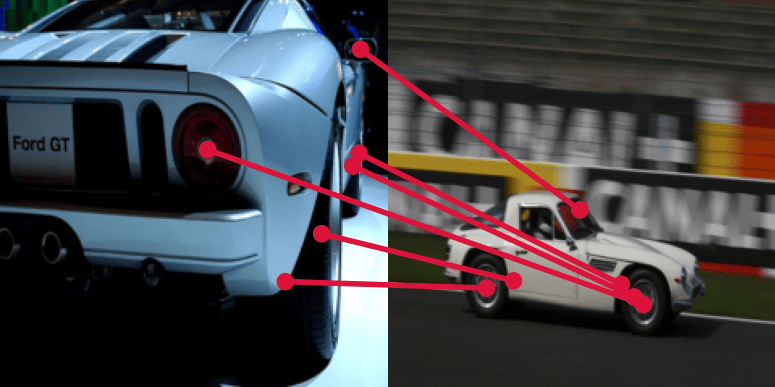}}\hfill
    \subfigure[]
	{\includegraphics[width=0.247\linewidth]{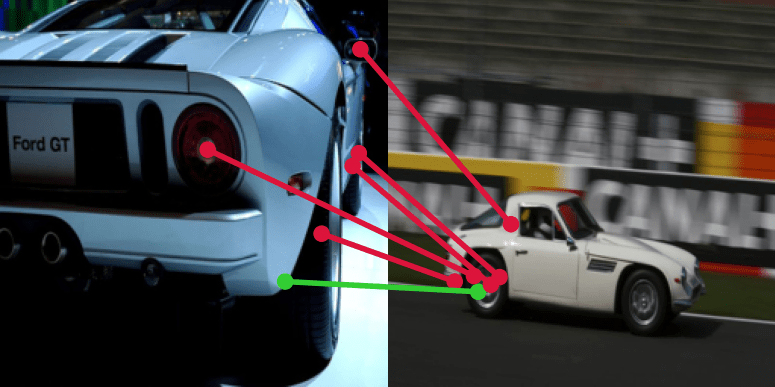}}\hfill
    \subfigure[]
	{\includegraphics[width=0.247\linewidth]{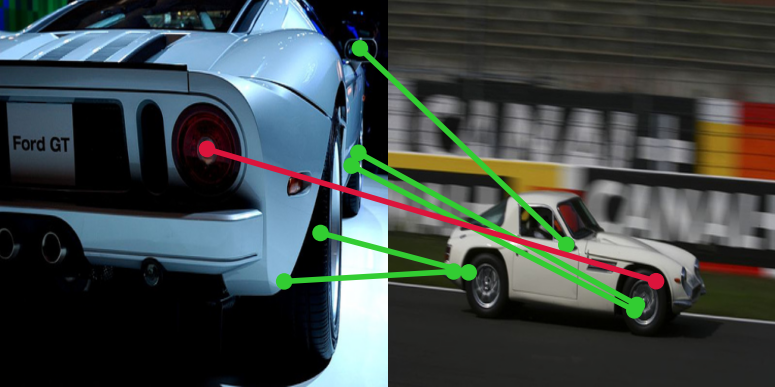}}\hfill\\
	\vspace{-20.5pt}
    \subfigure[]
	{\includegraphics[width=0.247\linewidth]{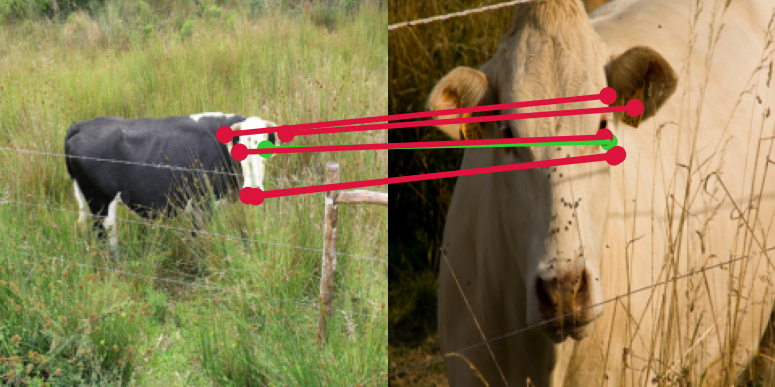}}\hfill
    \subfigure[]
	{\includegraphics[width=0.247\linewidth]{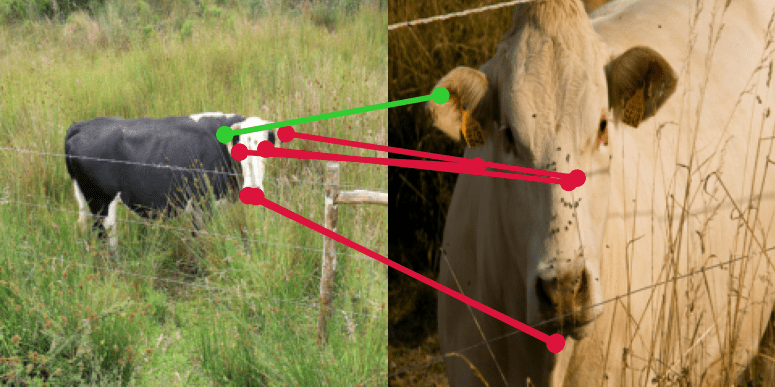}}\hfill
    \subfigure[]
	{\includegraphics[width=0.247\linewidth]{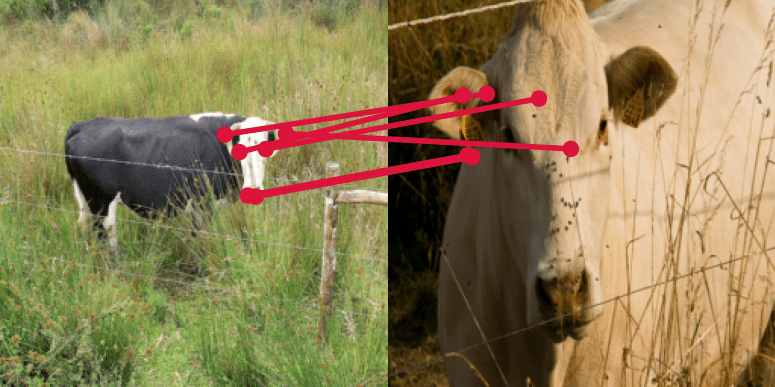}}\hfill
    \subfigure[]
	{\includegraphics[width=0.247\linewidth]{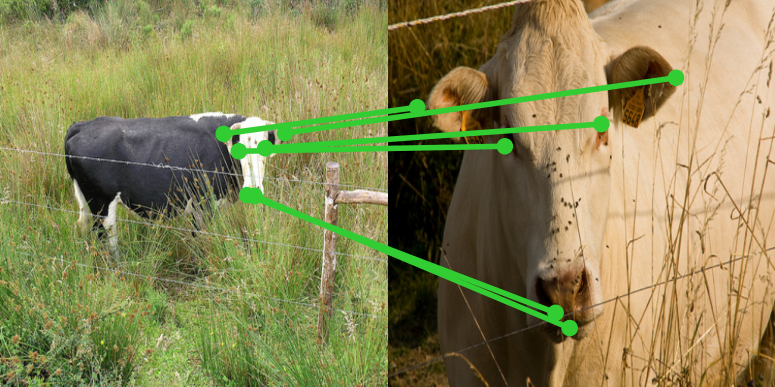}}\hfill\\
	\vspace{-20.5pt}
    \subfigure[]
	{\includegraphics[width=0.247\linewidth]{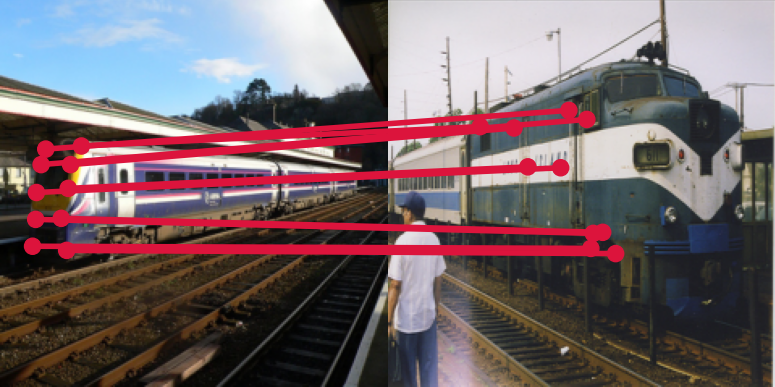}}\hfill
    \subfigure[]
	{\includegraphics[width=0.247\linewidth]{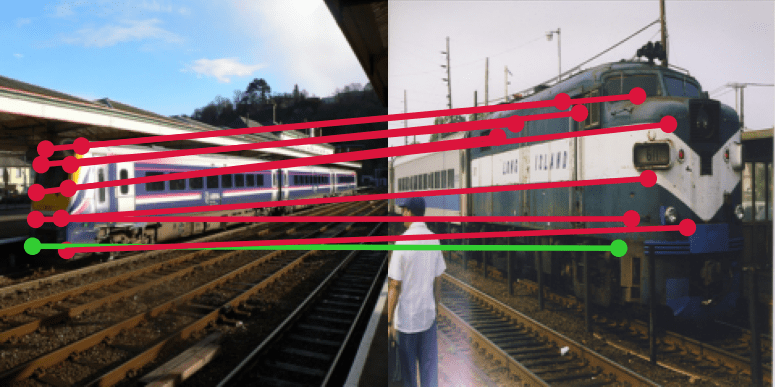}}\hfill
    \subfigure[]
	{\includegraphics[width=0.247\linewidth]{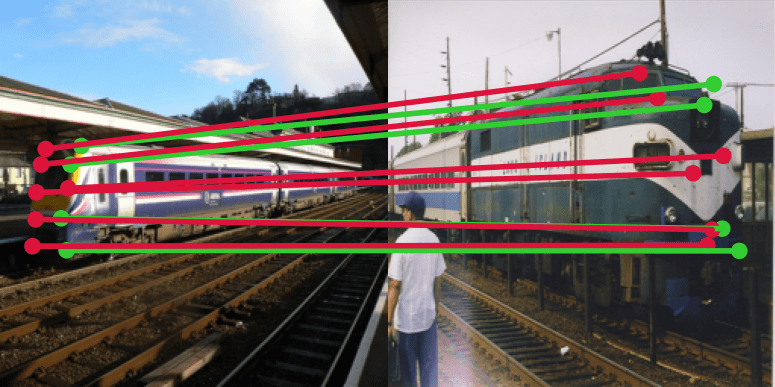}}\hfill
    \subfigure[]
	{\includegraphics[width=0.247\linewidth]{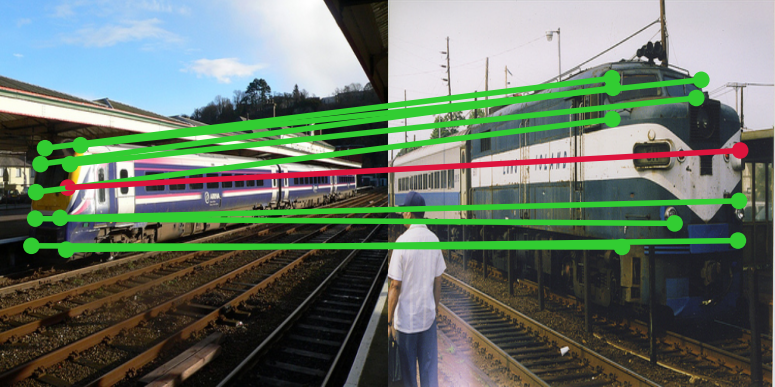}}\hfill\\
	\vspace{-20.5pt}
    \subfigure[]
	{\includegraphics[width=0.247\linewidth]{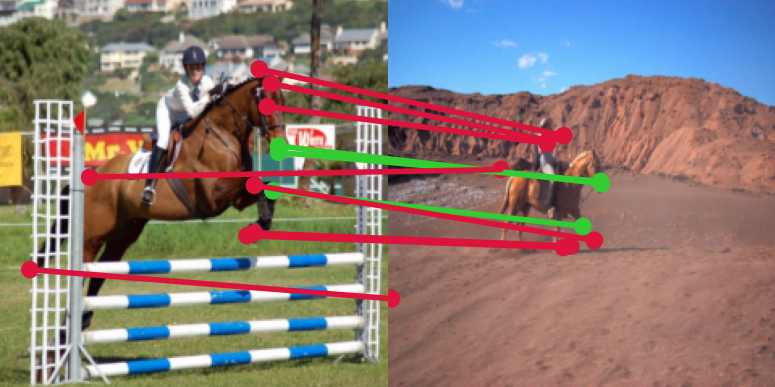}}\hfill
    \subfigure[]
	{\includegraphics[width=0.247\linewidth]{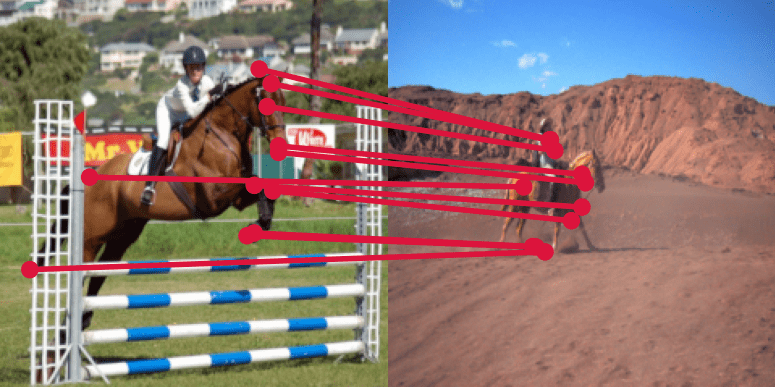}}\hfill
    \subfigure[]
	{\includegraphics[width=0.247\linewidth]{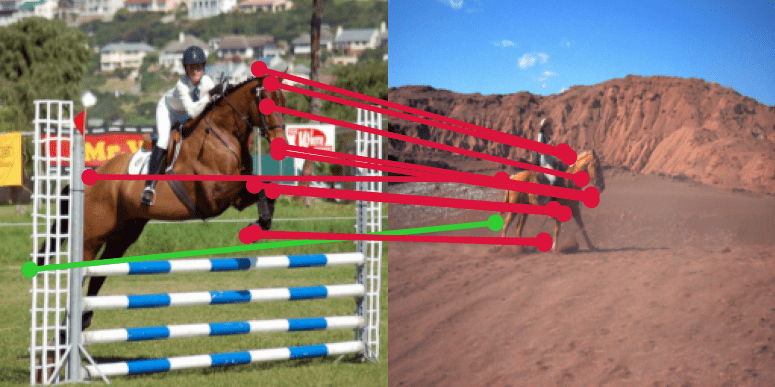}}\hfill
    \subfigure[]
	{\includegraphics[width=0.247\linewidth]{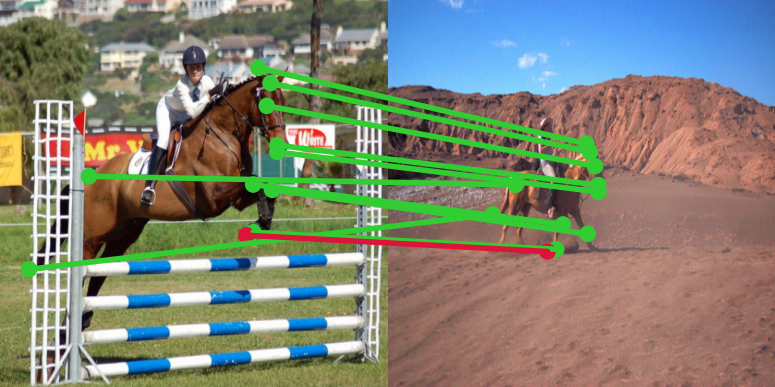}}\hfill\\
	\vspace{-20.5pt}
    \subfigure[]
	{\includegraphics[width=0.247\linewidth]{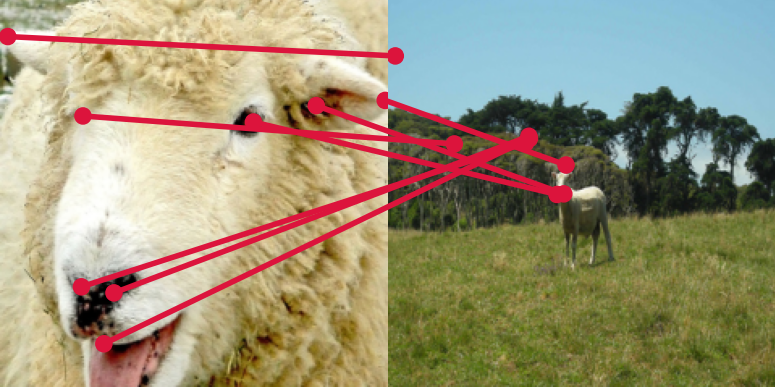}}\hfill
    \subfigure[]
	{\includegraphics[width=0.247\linewidth]{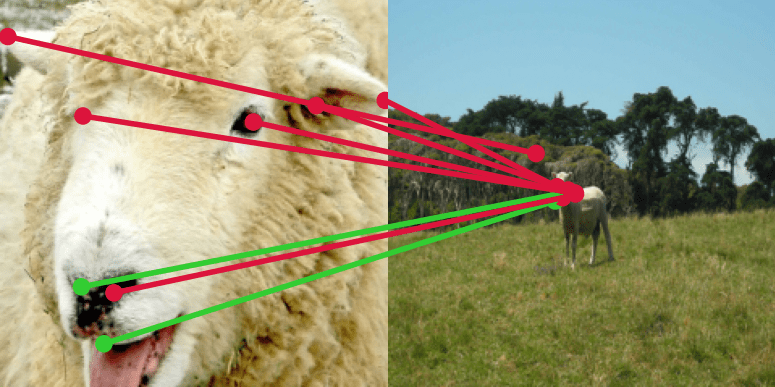}}\hfill
    \subfigure[]
	{\includegraphics[width=0.247\linewidth]{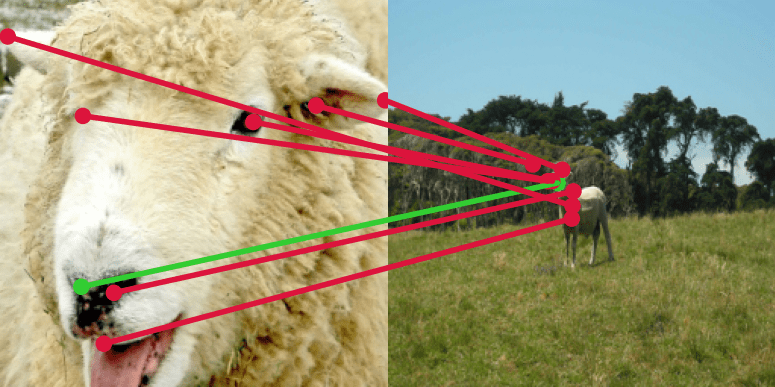}}\hfill
    \subfigure[]
	{\includegraphics[width=0.247\linewidth]{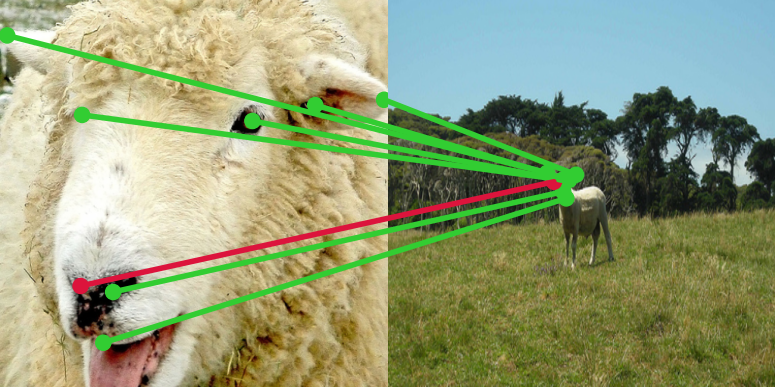}}\hfill\\
	\vspace{-20.5pt}
    \subfigure[]
	{\includegraphics[width=0.247\linewidth]{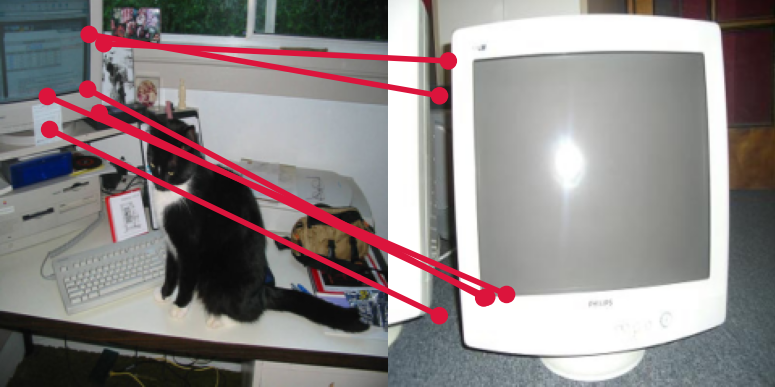}}\hfill
    \subfigure[]
	{\includegraphics[width=0.247\linewidth]{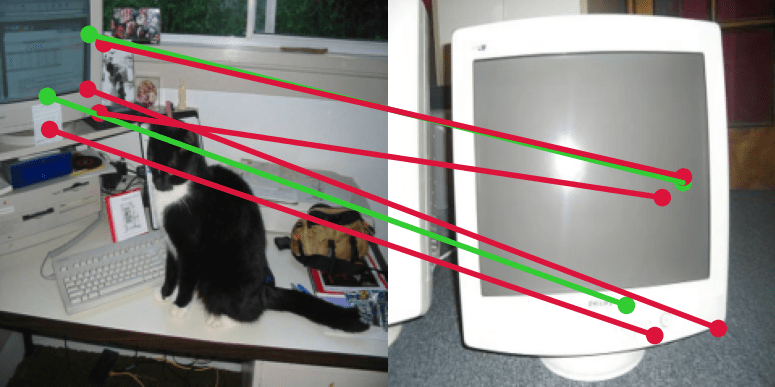}}\hfill
    \subfigure[]
	{\includegraphics[width=0.247\linewidth]{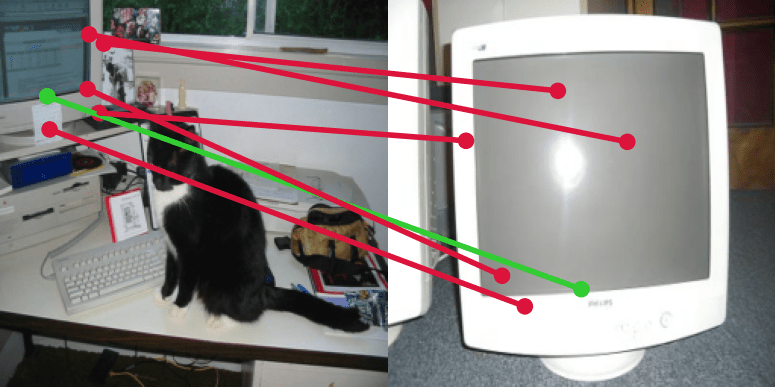}}\hfill
    \subfigure[]
	{\includegraphics[width=0.247\linewidth]{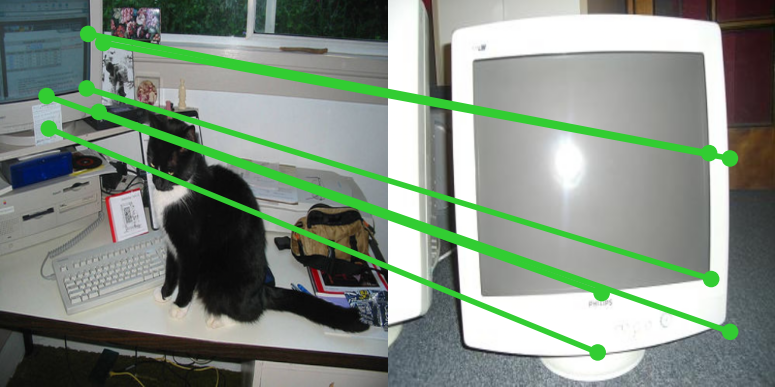}}\hfill\\
	\vspace{-20.5pt}
	\subfigure[(a) DHPF~\cite{min2020learning}]
	{\includegraphics[width=0.247\linewidth]{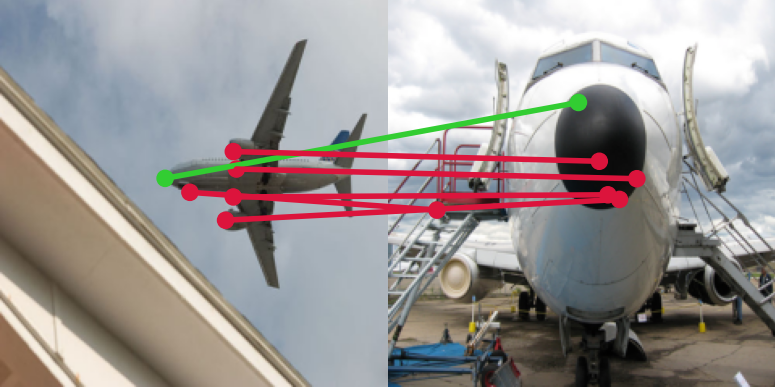}}\hfill
    \subfigure[(b) CHM~\cite{min2021convolutional}]
	{\includegraphics[width=0.247\linewidth]{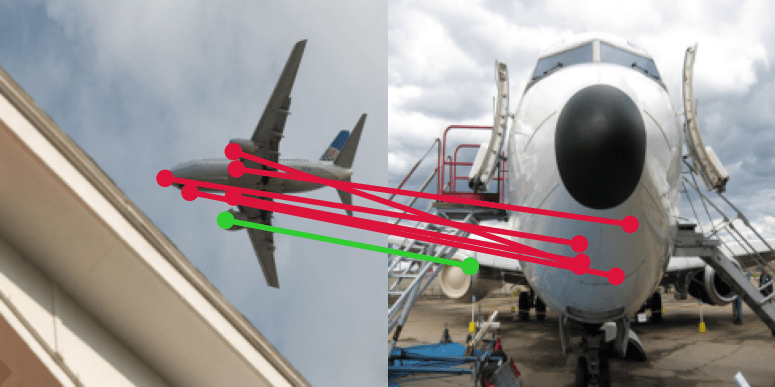}}\hfill
    \subfigure[(c) CATs~\cite{cho2021semantic}]
	{\includegraphics[width=0.247\linewidth]{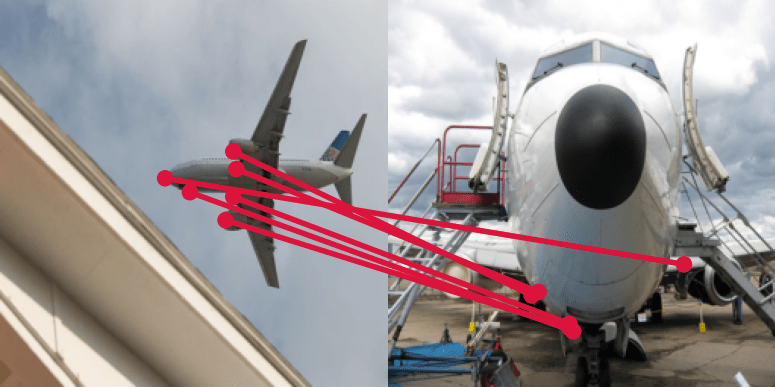}}\hfill
    \subfigure[(d) IFCAT]
	{\includegraphics[width=0.247\linewidth]{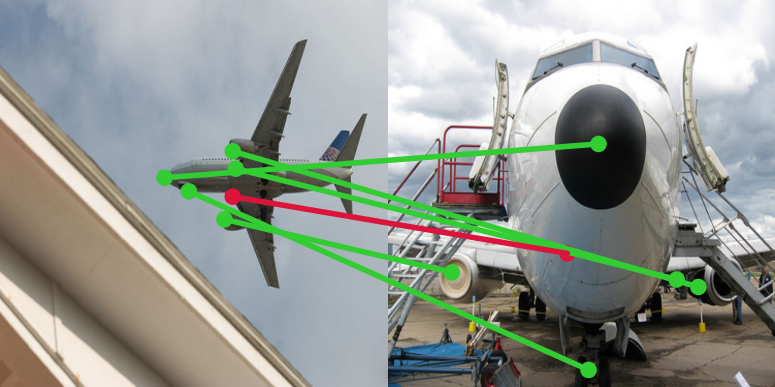}}\hfill\\
	\vspace{-5pt}
    \caption{\textbf{Qualitative results on SPair-71k~\cite{min2019spair}:} keypoints transfer results by (a) DHPF~\cite{min2020learning}, (b) CHM~\cite{min2021convolutional}, and (c) CATs~\cite{cho2021semantic}, and (d) ours. Note that green and red lines denote correct and wrong predictions, respectively, with respect to the ground-truth.}\label{fig:spair_quali}\vspace{-10pt}
  
\end{figure}
\begin{figure}[!t]
    \centering
    \renewcommand{\thesubfigure}{}
    \subfigure[]
	{\includegraphics[width=0.247\linewidth]{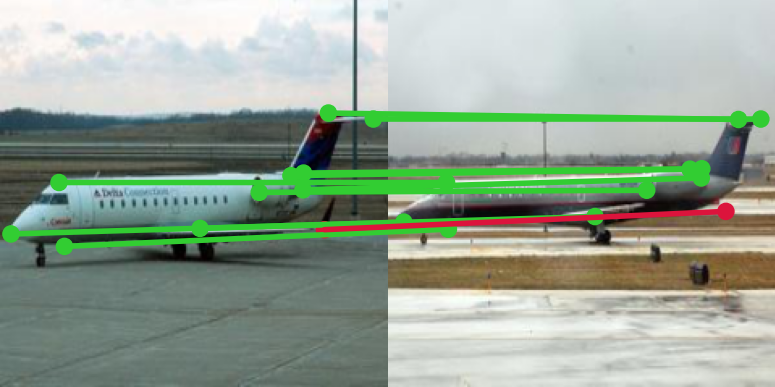}}\hfill
    \subfigure[]
	{\includegraphics[width=0.247\linewidth]{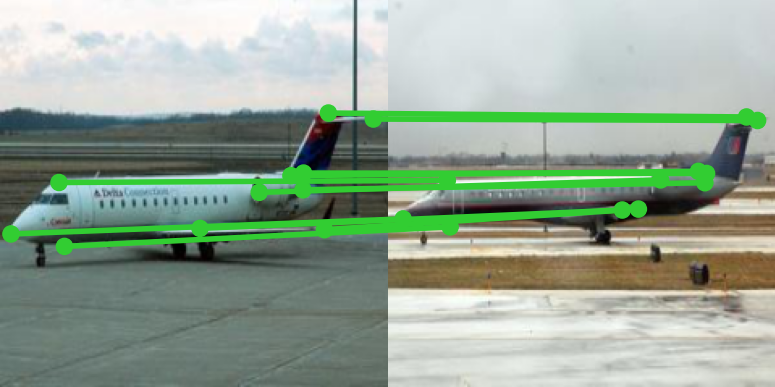}}\hfill
    \subfigure[]
	{\includegraphics[width=0.247\linewidth]{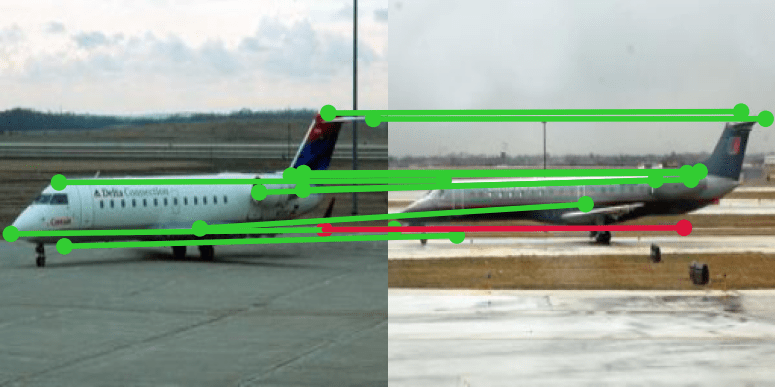}}\hfill
    \subfigure[]
	{\includegraphics[width=0.247\linewidth]{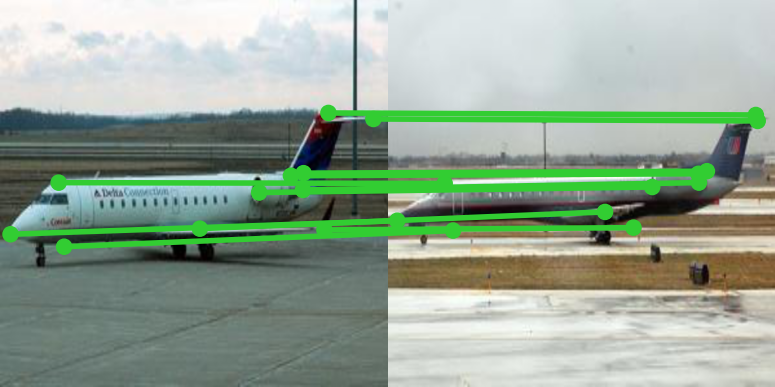}}\hfill\\
	\vspace{-20.5pt}
    \subfigure[]
	{\includegraphics[width=0.247\linewidth]{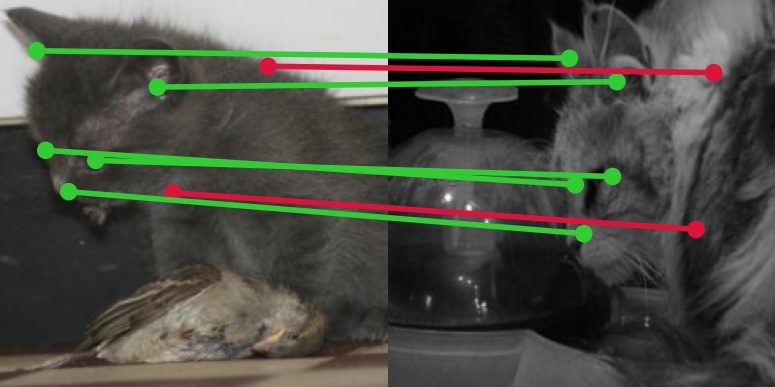}}\hfill
    \subfigure[]
	{\includegraphics[width=0.247\linewidth]{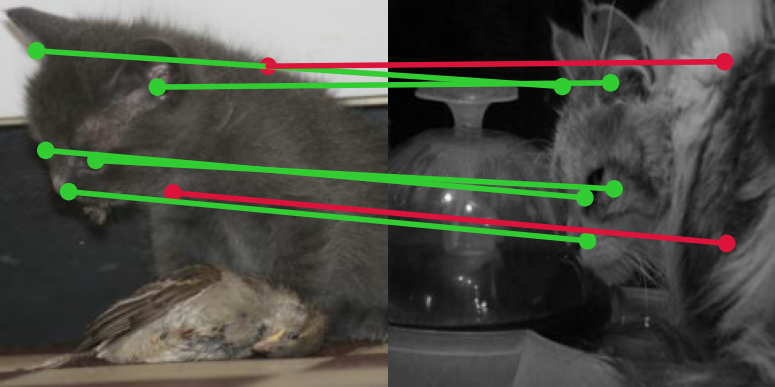}}\hfill
    \subfigure[]
	{\includegraphics[width=0.247\linewidth]{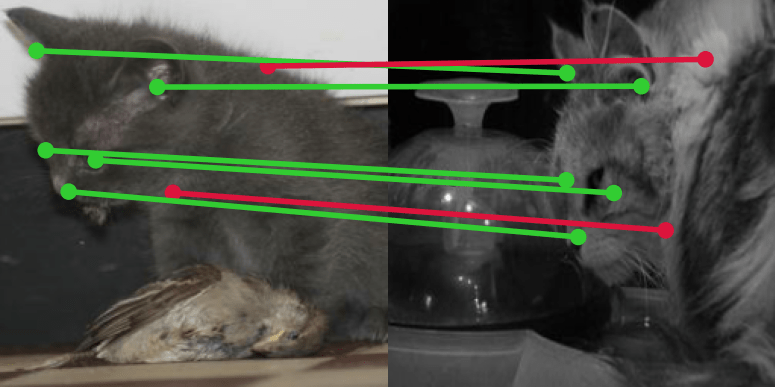}}\hfill
    \subfigure[]
	{\includegraphics[width=0.247\linewidth]{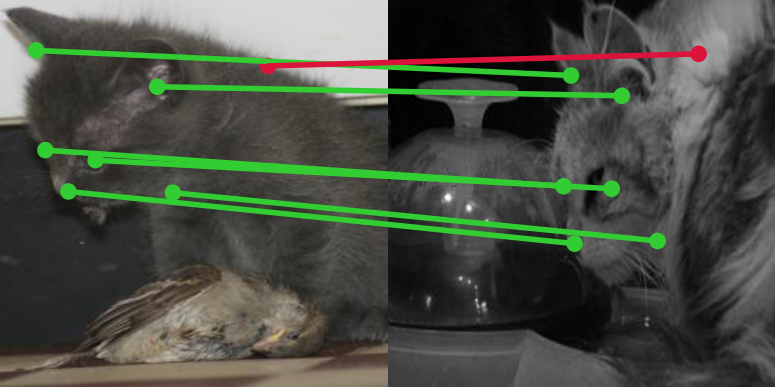}}\hfill\\
	\vspace{-20.5pt}
    \subfigure[]
	{\includegraphics[width=0.247\linewidth]{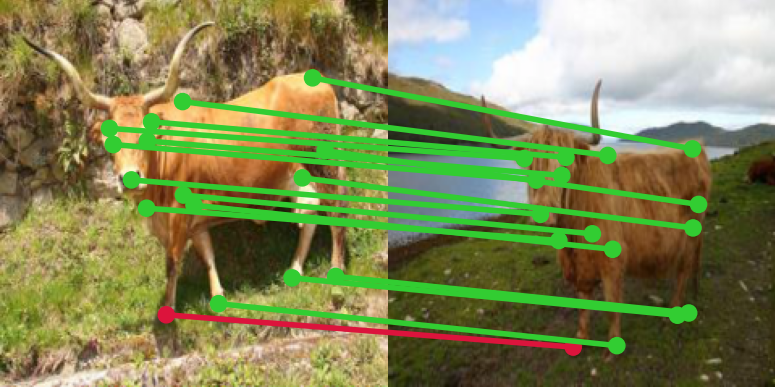}}\hfill
    \subfigure[]
	{\includegraphics[width=0.247\linewidth]{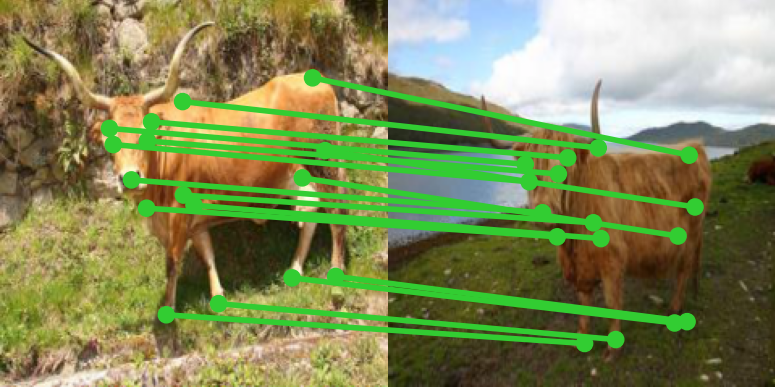}}\hfill
    \subfigure[]
	{\includegraphics[width=0.247\linewidth]{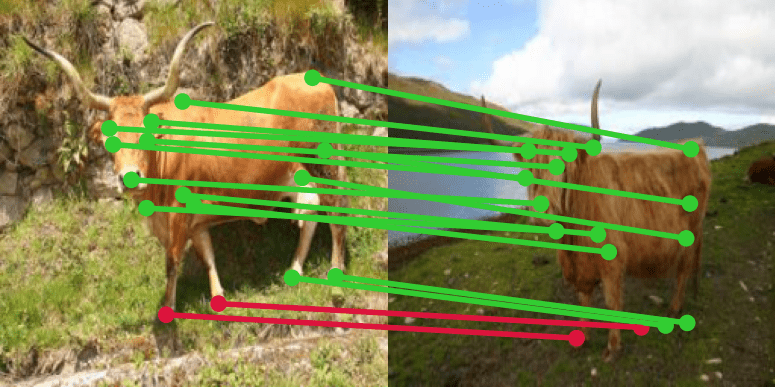}}\hfill
    \subfigure[]
	{\includegraphics[width=0.247\linewidth]{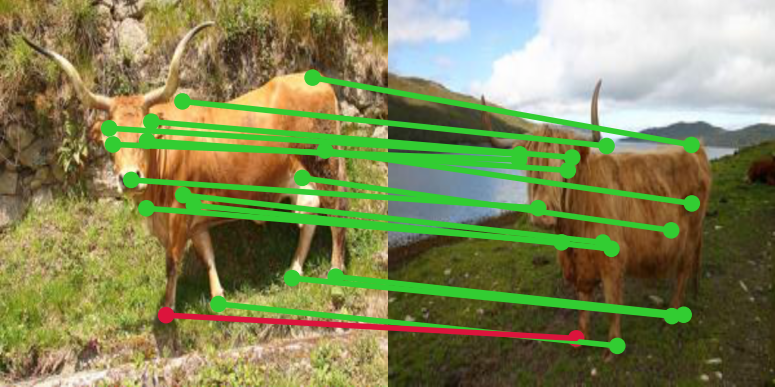}}\hfill\\
	\vspace{-20.5pt}
    \subfigure[]
	{\includegraphics[width=0.247\linewidth]{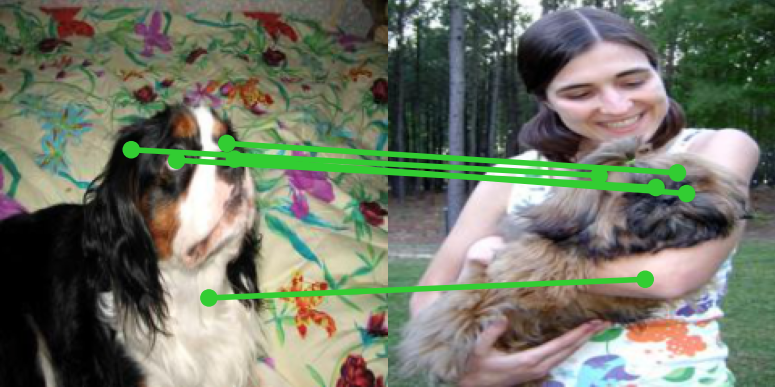}}\hfill
    \subfigure[]
	{\includegraphics[width=0.247\linewidth]{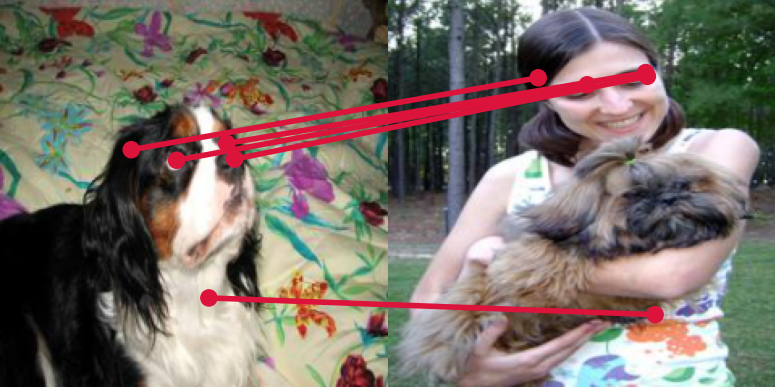}}\hfill
    \subfigure[]
	{\includegraphics[width=0.247\linewidth]{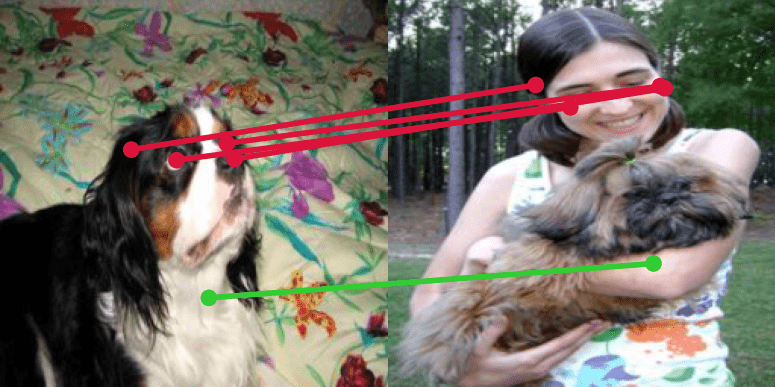}}\hfill
    \subfigure[]
	{\includegraphics[width=0.247\linewidth]{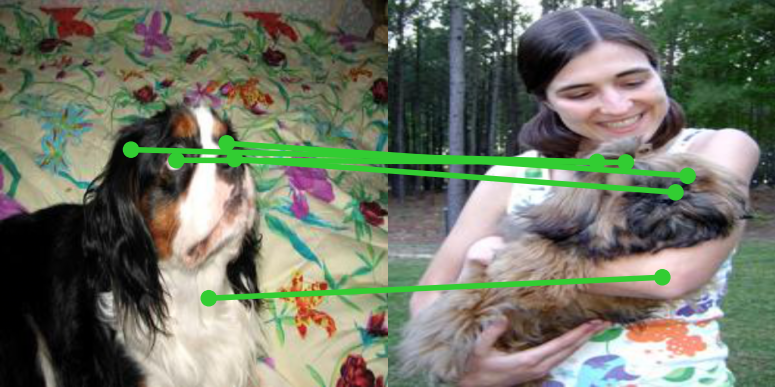}}\hfill\\
	\vspace{-20.5pt}
	\subfigure[(a) DHPF~\cite{min2020learning}]
	{\includegraphics[width=0.247\linewidth]{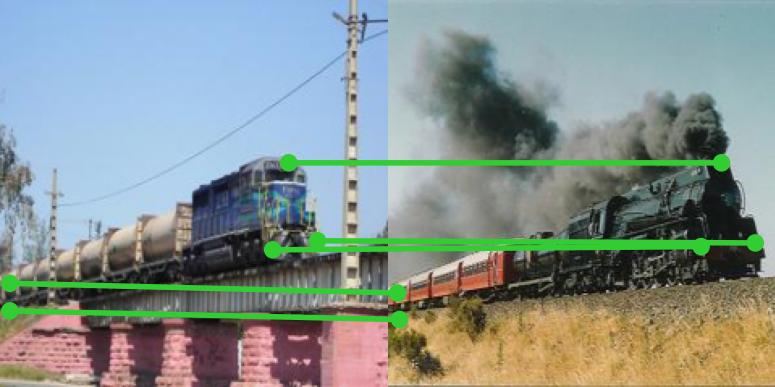}}\hfill
    \subfigure[(b) CHM~\cite{min2021convolutional}]
	{\includegraphics[width=0.247\linewidth]{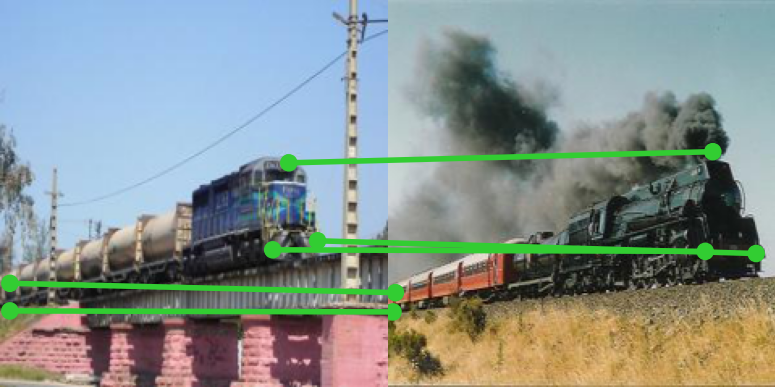}}\hfill
    \subfigure[(c) CATs~\cite{cho2021semantic}]
	{\includegraphics[width=0.247\linewidth]{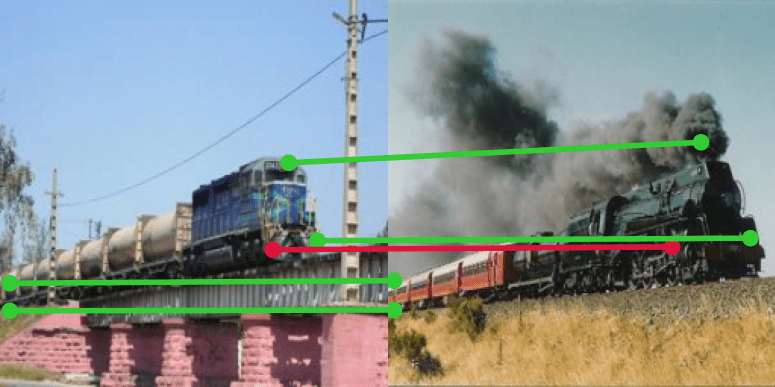}}\hfill
    \subfigure[(d) IFCAT]
	{\includegraphics[width=0.247\linewidth]{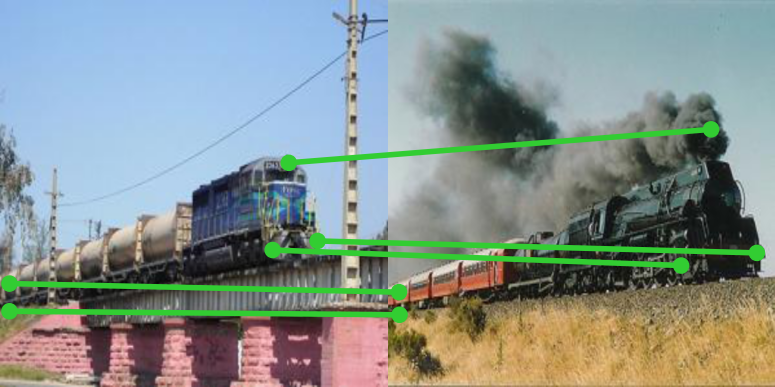}}\hfill\\
	\vspace{-5pt}
    \caption{\textbf{Qualitative results on PF-PASCAL~\cite{ham2017proposal}} }\label{fig:pascal_quali}\vspace{-10pt}
\end{figure}
\begin{figure}[!t]
    \centering
    \renewcommand{\thesubfigure}{}
    \subfigure[]
	{\includegraphics[width=0.247\linewidth]{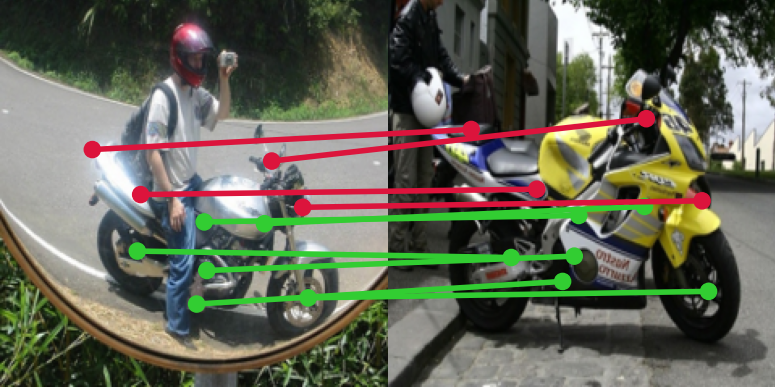}}\hfill
    \subfigure[]
	{\includegraphics[width=0.247\linewidth]{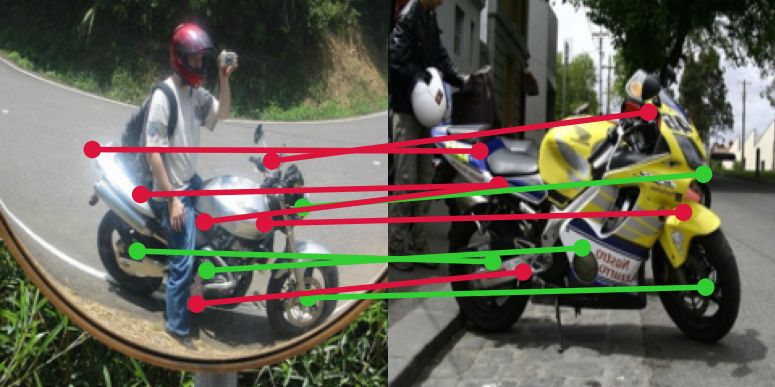}}\hfill
    \subfigure[]
	{\includegraphics[width=0.247\linewidth]{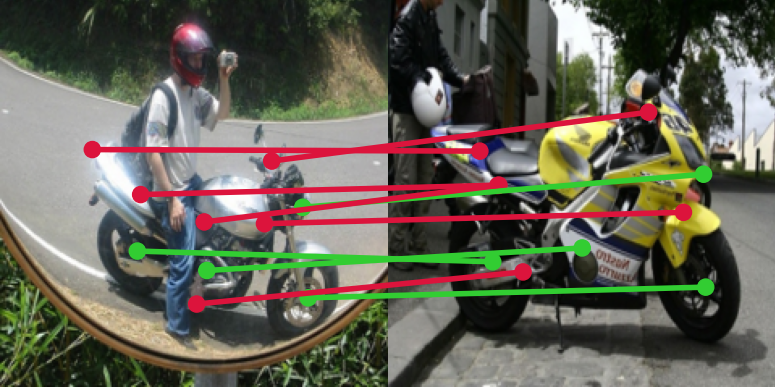}}\hfill
    \subfigure[]
	{\includegraphics[width=0.247\linewidth]{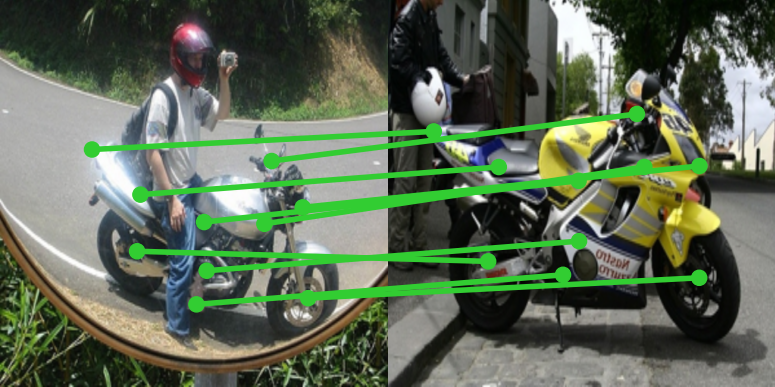}}\hfill\\
	\vspace{-20.5pt}
    \subfigure[]
	{\includegraphics[width=0.247\linewidth]{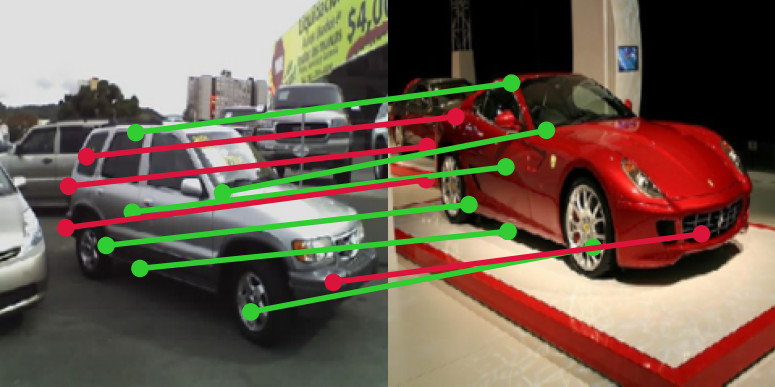}}\hfill
    \subfigure[]
	{\includegraphics[width=0.247\linewidth]{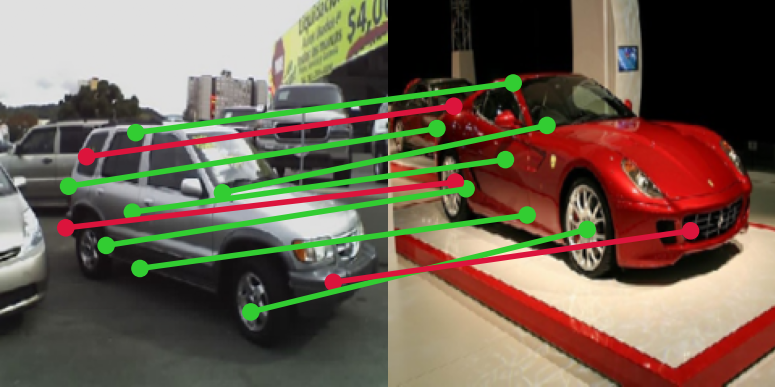}}\hfill
    \subfigure[]
	{\includegraphics[width=0.247\linewidth]{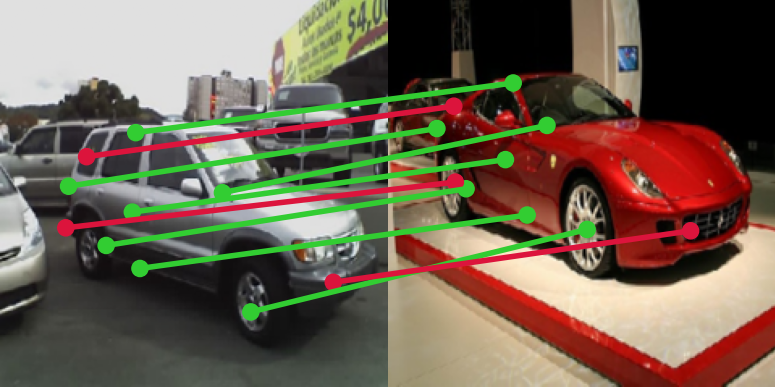}}\hfill
    \subfigure[]
	{\includegraphics[width=0.247\linewidth]{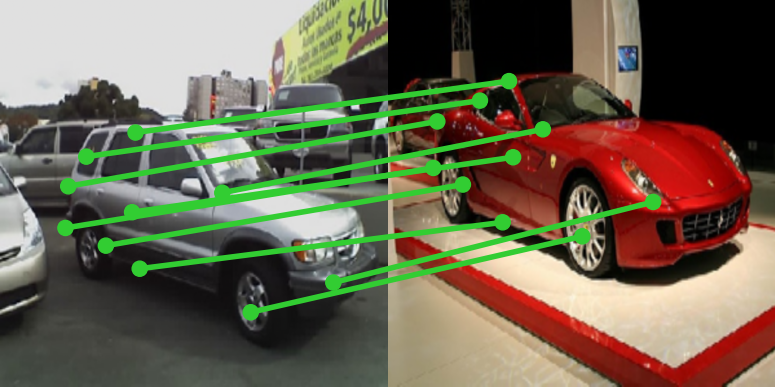}}\hfill\\
	\vspace{-20.5pt}
    \subfigure[]
	{\includegraphics[width=0.247\linewidth]{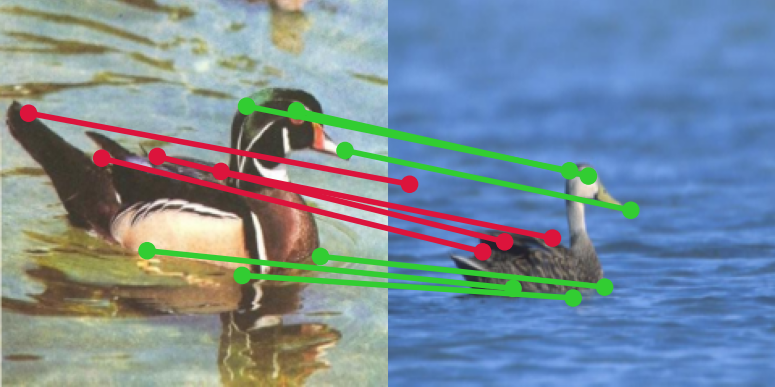}}\hfill
    \subfigure[]
	{\includegraphics[width=0.247\linewidth]{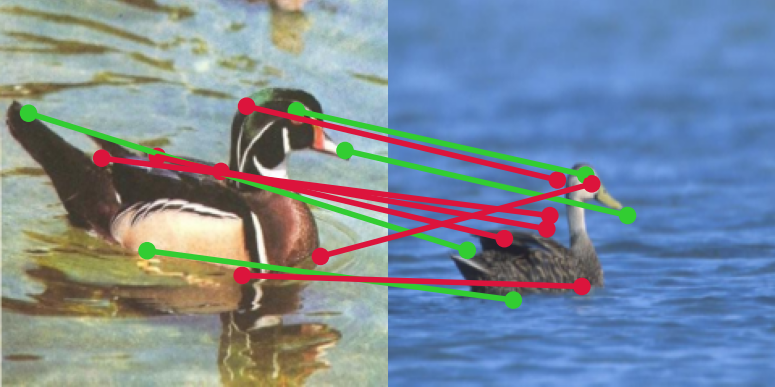}}\hfill
    \subfigure[]
	{\includegraphics[width=0.247\linewidth]{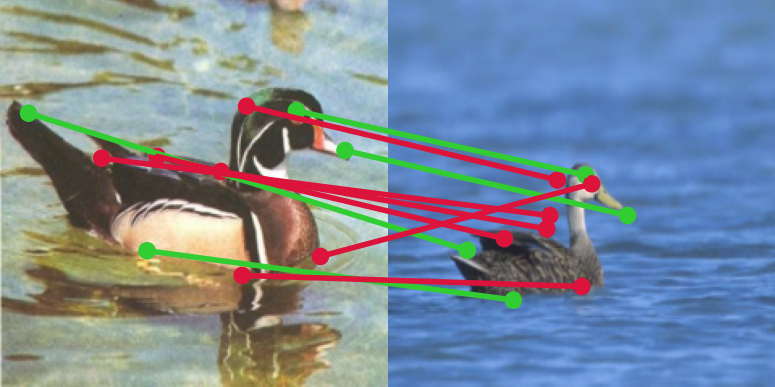}}\hfill
    \subfigure[]
	{\includegraphics[width=0.247\linewidth]{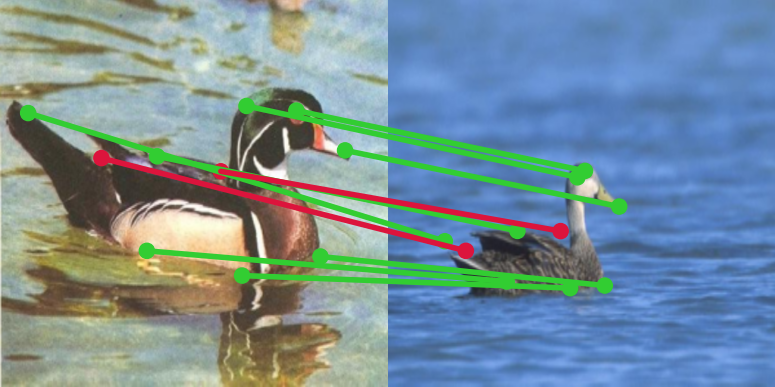}}\hfill\\
	\vspace{-20.5pt}
    \subfigure[]
	{\includegraphics[width=0.247\linewidth]{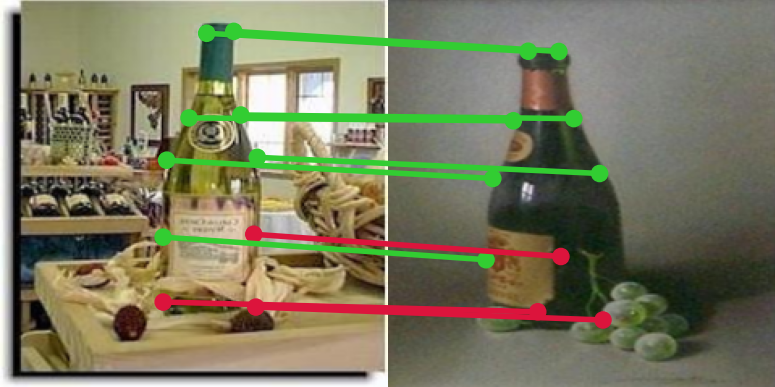}}\hfill
    \subfigure[]
	{\includegraphics[width=0.247\linewidth]{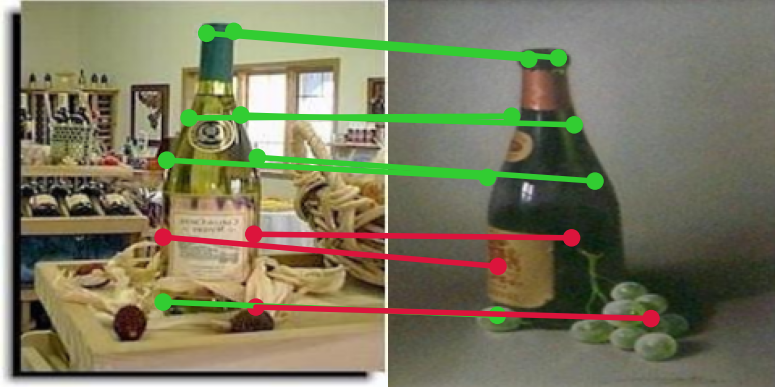}}\hfill
    \subfigure[]
	{\includegraphics[width=0.247\linewidth]{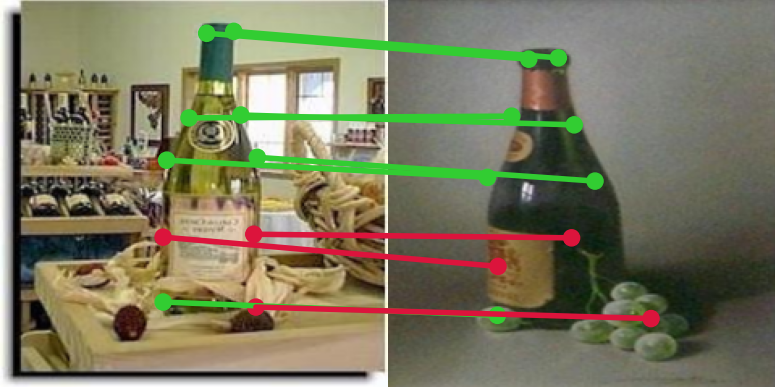}}\hfill
    \subfigure[]
	{\includegraphics[width=0.247\linewidth]{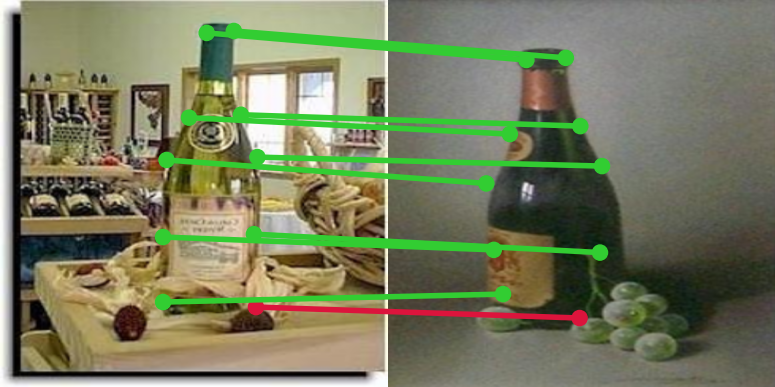}}\hfill\\
	\vspace{-20.5pt}
	\subfigure[(a) DHPF~\cite{min2020learning}]
	{\includegraphics[width=0.247\linewidth]{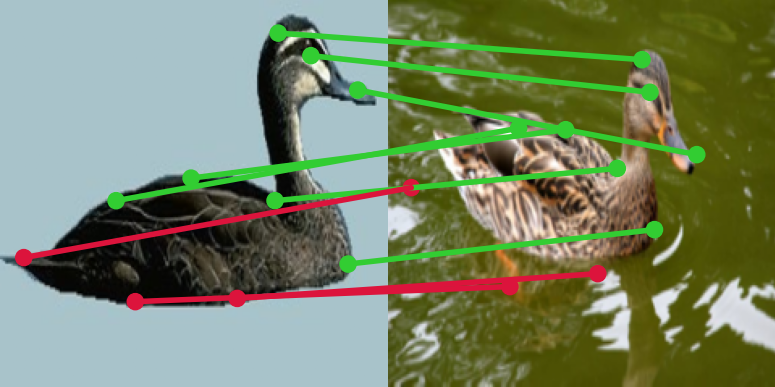}}\hfill
    \subfigure[(b) CHM~\cite{min2021convolutional}]
	{\includegraphics[width=0.247\linewidth]{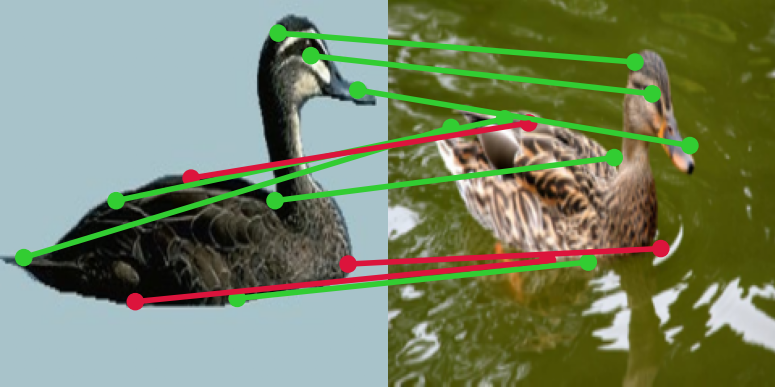}}\hfill
    \subfigure[(c) CATs~\cite{cho2021semantic}]
	{\includegraphics[width=0.247\linewidth]{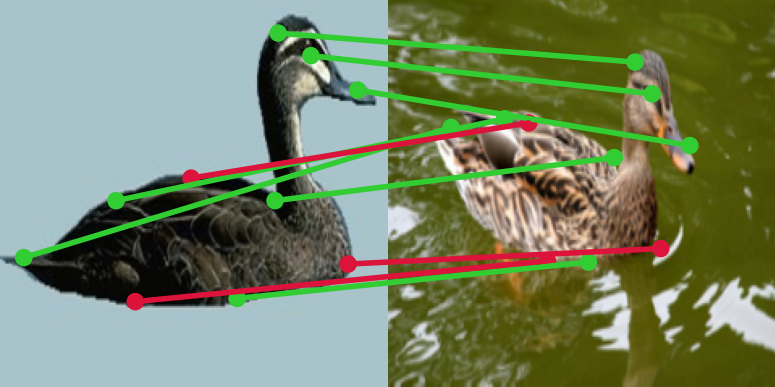}}\hfill
    \subfigure[(d) IFCAT]
	{\includegraphics[width=0.247\linewidth]{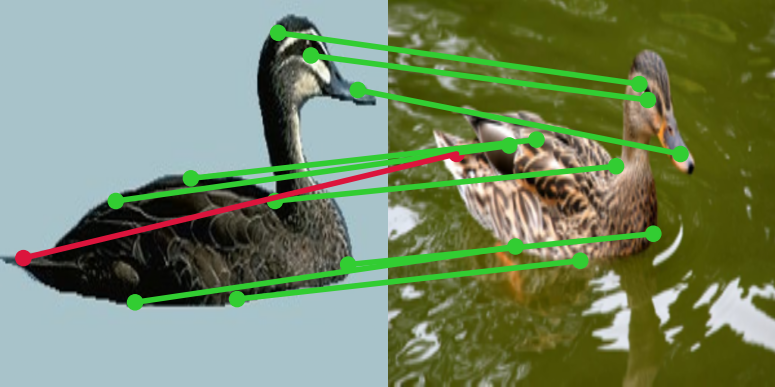}}\hfill\\
	
	\vspace{-5pt}
    \caption{\textbf{Qualitative results on PF-WILLOW~\cite{ham2016proposal}.} }\label{fig:willow_quali}\vspace{-10pt}
\end{figure}
\begin{figure}[!t]
    \centering
    \renewcommand{\thesubfigure}{}
    \subfigure[]
	{\includegraphics[width=0.247\linewidth]{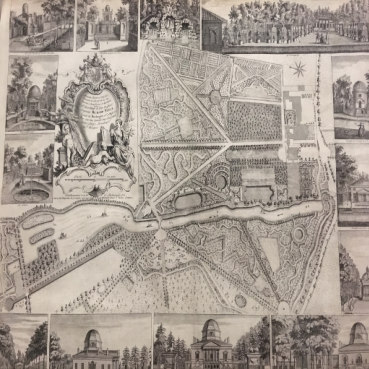}}\hfill
    \subfigure[]
	{\includegraphics[width=0.247\linewidth]{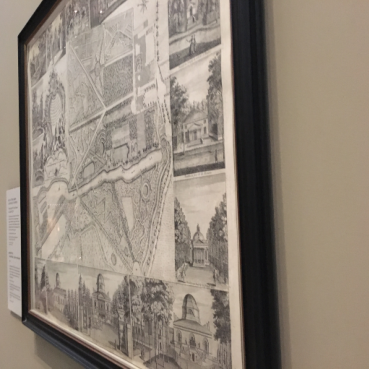}}\hfill
    \subfigure[]
	{\includegraphics[width=0.247\linewidth]{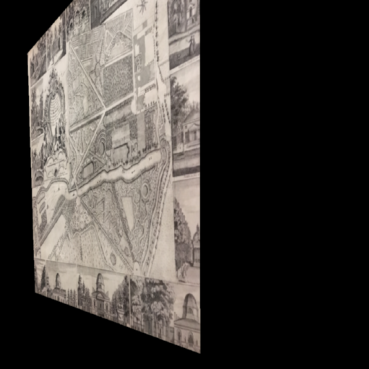}}\hfill
    \subfigure[]
	{\includegraphics[width=0.247\linewidth]{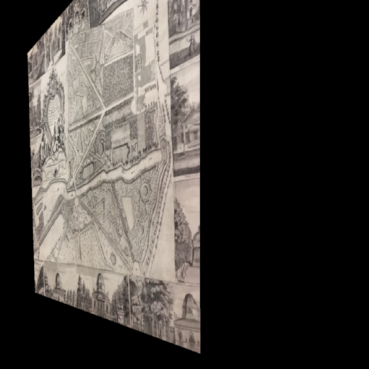}}\hfill\\
	\vspace{-20.5pt}
    \subfigure[]
	{\includegraphics[width=0.247\linewidth]{figure/supple_qual/hpatches/03_src_img.png}}\hfill
    \subfigure[]
	{\includegraphics[width=0.247\linewidth]{figure/supple_qual/hpatches/03_trg_img.png}}\hfill
    \subfigure[]
	{\includegraphics[width=0.247\linewidth]{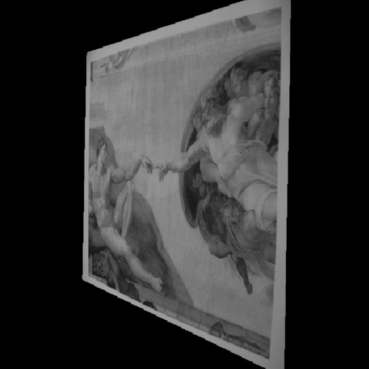}}\hfill
    \subfigure[]
	{\includegraphics[width=0.247\linewidth]{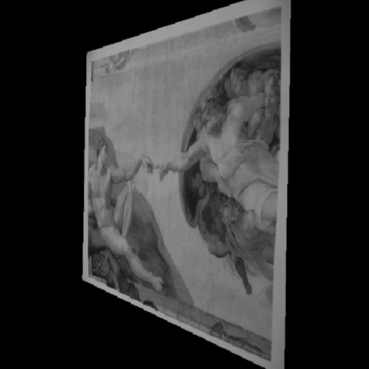}}\hfill\\
	\vspace{-20.5pt}
    \subfigure[]
	{\includegraphics[width=0.247\linewidth]{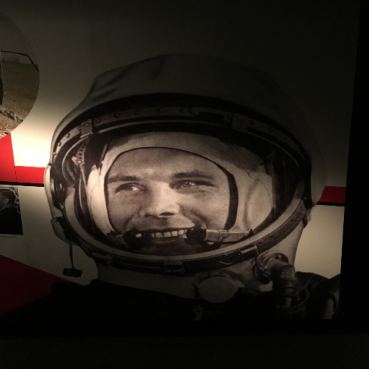}}\hfill
    \subfigure[]
	{\includegraphics[width=0.247\linewidth]{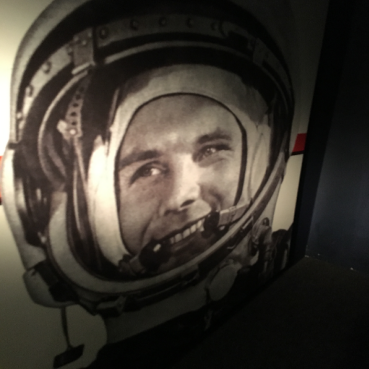}}\hfill
    \subfigure[]
	{\includegraphics[width=0.247\linewidth]{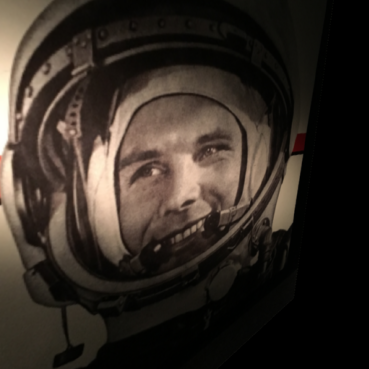}}\hfill
    \subfigure[]
	{\includegraphics[width=0.247\linewidth]{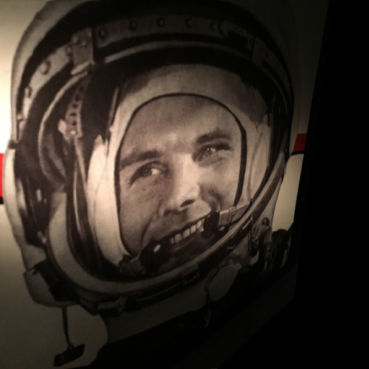}}\hfill\\
	\vspace{-20.5pt}
    \subfigure[]
	{\includegraphics[width=0.247\linewidth]{figure/supple_qual/hpatches/15_src_img.png}}\hfill
    \subfigure[]
	{\includegraphics[width=0.247\linewidth]{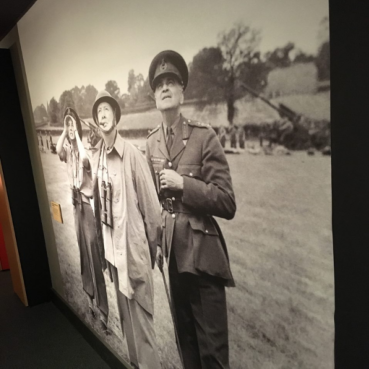}}\hfill
    \subfigure[]
	{\includegraphics[width=0.247\linewidth]{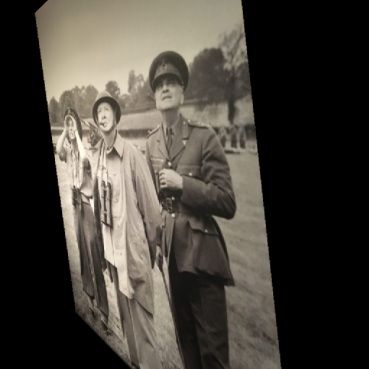}}\hfill
    \subfigure[]
	{\includegraphics[width=0.247\linewidth]{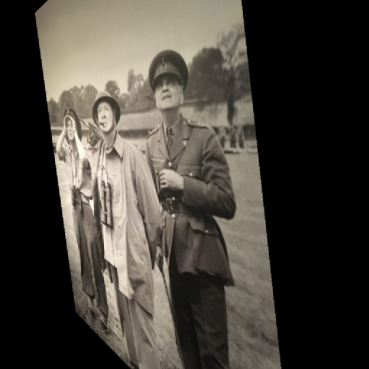}}\hfill\\
	\vspace{-20.5pt}
    \subfigure[]
	{\includegraphics[width=0.247\linewidth]{figure/supple_qual/hpatches/24_src_img.png}}\hfill
    \subfigure[]
	{\includegraphics[width=0.247\linewidth]{figure/supple_qual/hpatches/24_trg_img.png}}\hfill
    \subfigure[]
	{\includegraphics[width=0.247\linewidth]{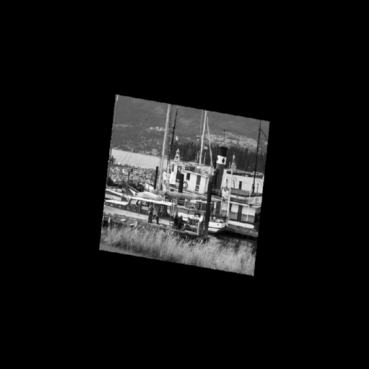}}\hfill
    \subfigure[]
	{\includegraphics[width=0.247\linewidth]{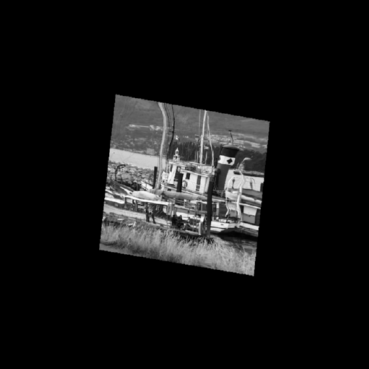}}\hfill\\
	\vspace{-20.5pt}
	\subfigure[Source]
	{\includegraphics[width=0.247\linewidth]{figure/supple_qual/hpatches/55_src_img.png}}\hfill
    \subfigure[Target]
	{\includegraphics[width=0.247\linewidth]{figure/supple_qual/hpatches/55_trg_img.png}}\hfill
    \subfigure[GT]
	{\includegraphics[width=0.247\linewidth]{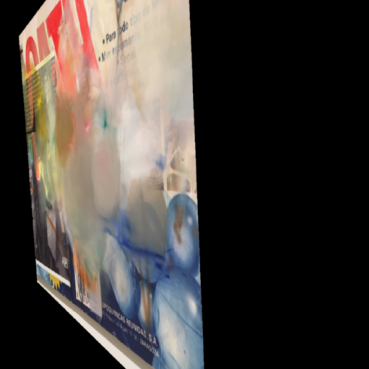}}\hfill
    \subfigure[Prediction]
	{\includegraphics[width=0.247\linewidth]{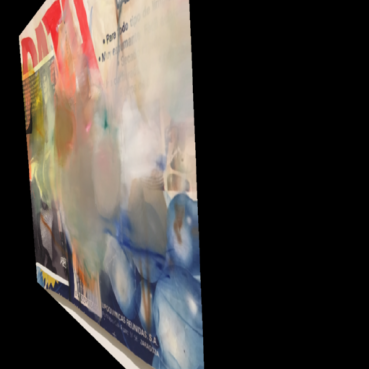}}\hfill\\
	
	\vspace{-5pt}
    \caption{\textbf{Qualitative results on HPatches~\cite{balntas2017hpatches}.} }\label{hpatches}\vspace{-10pt}
\end{figure}
\begin{figure}[!t]
    \centering
    \renewcommand{\thesubfigure}{}
    \subfigure[]
	{\includegraphics[width=0.123\linewidth]{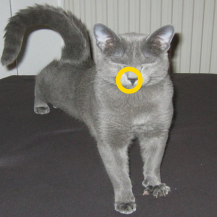}}\hfill
    \subfigure[]
	{\includegraphics[width=0.123\linewidth]{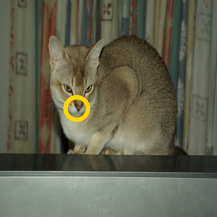}}\hfill
    \subfigure[]
	{\includegraphics[width=0.123\linewidth]{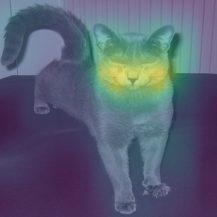}}\hfill
    \subfigure[]
	{\includegraphics[width=0.123\linewidth]{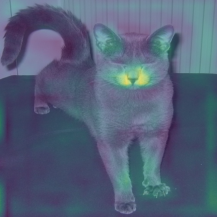}}\hfill
    \subfigure[]
	{\includegraphics[width=0.123\linewidth]{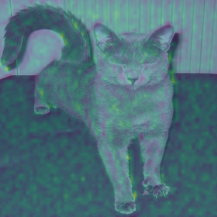}}\hfill
    \subfigure[]
	{\includegraphics[width=0.123\linewidth]{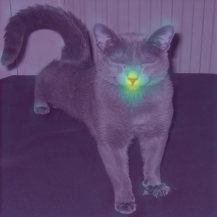}}\hfill
    \subfigure[]
	{\includegraphics[width=0.123\linewidth]{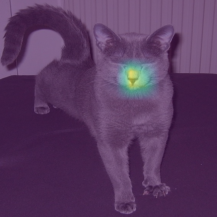}}\hfill
    \subfigure[]
	{\includegraphics[width=0.123\linewidth]{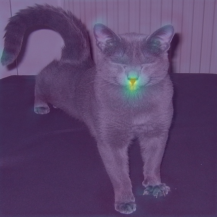}}\hfill\\
	\vspace{-20.5pt}
    \subfigure[]
	{\includegraphics[width=0.123\linewidth]{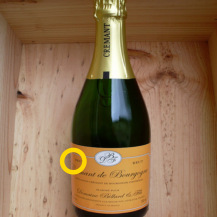}}\hfill
    \subfigure[]
	{\includegraphics[width=0.123\linewidth]{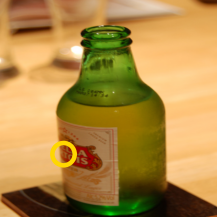}}\hfill
    \subfigure[]
	{\includegraphics[width=0.123\linewidth]{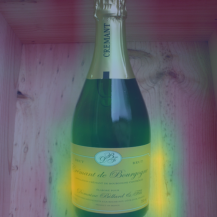}}\hfill
    \subfigure[]
	{\includegraphics[width=0.123\linewidth]{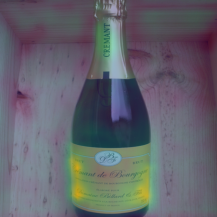}}\hfill
    \subfigure[]
	{\includegraphics[width=0.123\linewidth]{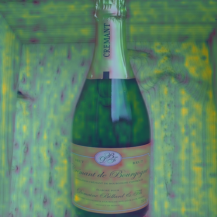}}\hfill
    \subfigure[]
	{\includegraphics[width=0.123\linewidth]{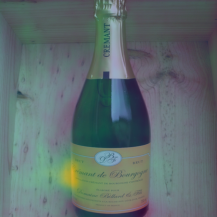}}\hfill
    \subfigure[]
	{\includegraphics[width=0.123\linewidth]{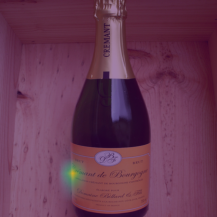}}\hfill
    \subfigure[]
	{\includegraphics[width=0.123\linewidth]{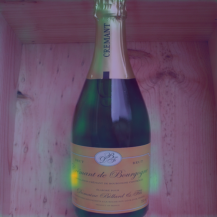}}\hfill\\
	\vspace{-20.5pt}
    \subfigure[]
	{\includegraphics[width=0.123\linewidth]{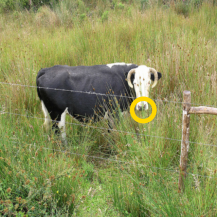}}\hfill
    \subfigure[]
	{\includegraphics[width=0.123\linewidth]{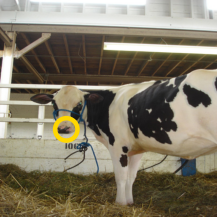}}\hfill
    \subfigure[]
	{\includegraphics[width=0.123\linewidth]{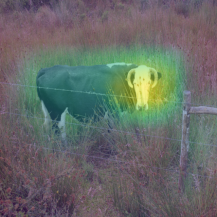}}\hfill
    \subfigure[]
	{\includegraphics[width=0.123\linewidth]{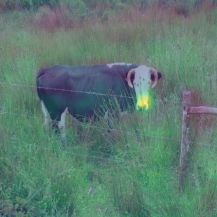}}\hfill
    \subfigure[]
	{\includegraphics[width=0.123\linewidth]{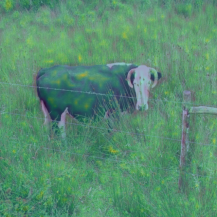}}\hfill
    \subfigure[]
	{\includegraphics[width=0.123\linewidth]{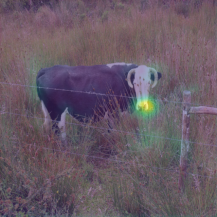}}\hfill
    \subfigure[]
	{\includegraphics[width=0.123\linewidth]{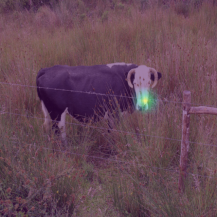}}\hfill
    \subfigure[]
	{\includegraphics[width=0.123\linewidth]{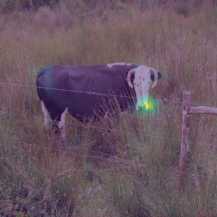}}\hfill\\
	\vspace{-20.5pt}
    \subfigure[]
	{\includegraphics[width=0.123\linewidth]{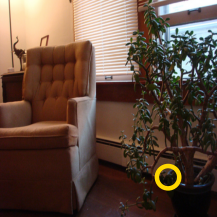}}\hfill
    \subfigure[]
	{\includegraphics[width=0.123\linewidth]{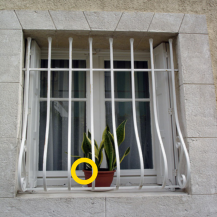}}\hfill
    \subfigure[]
	{\includegraphics[width=0.123\linewidth]{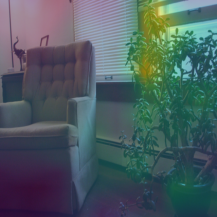}}\hfill
    \subfigure[]
	{\includegraphics[width=0.123\linewidth]{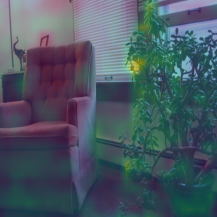}}\hfill
    \subfigure[]
	{\includegraphics[width=0.123\linewidth]{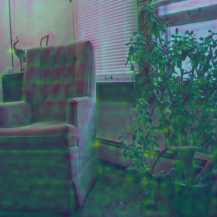}}\hfill
    \subfigure[]
	{\includegraphics[width=0.123\linewidth]{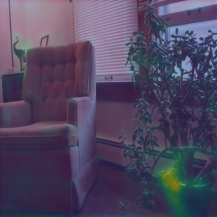}}\hfill
    \subfigure[]
	{\includegraphics[width=0.123\linewidth]{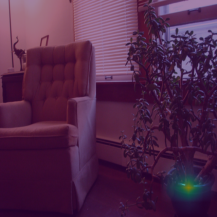}}\hfill
    \subfigure[]
	{\includegraphics[width=0.123\linewidth]{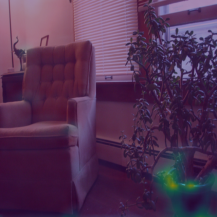}}\hfill\\
	\vspace{-20.5pt}
    \subfigure[]
	{\includegraphics[width=0.123\linewidth]{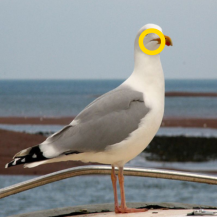}}\hfill
    \subfigure[]
	{\includegraphics[width=0.123\linewidth]{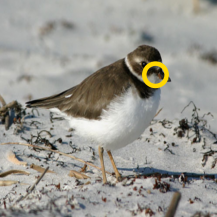}}\hfill
    \subfigure[]
	{\includegraphics[width=0.123\linewidth]{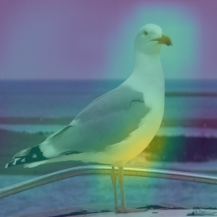}}\hfill
    \subfigure[]
	{\includegraphics[width=0.123\linewidth]{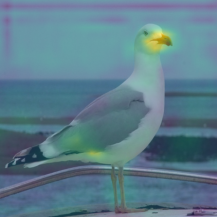}}\hfill
    \subfigure[]
	{\includegraphics[width=0.123\linewidth]{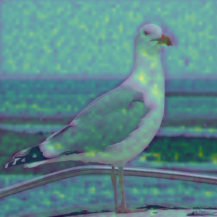}}\hfill
    \subfigure[]
	{\includegraphics[width=0.123\linewidth]{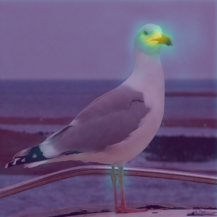}}\hfill
    \subfigure[]
	{\includegraphics[width=0.123\linewidth]{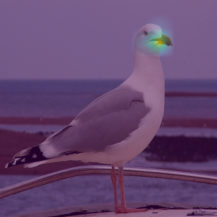}}\hfill
    \subfigure[]
	{\includegraphics[width=0.123\linewidth]{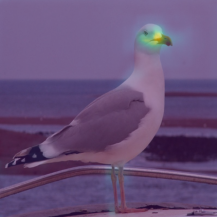}}\hfill\\
	\vspace{-20.5pt}
    \subfigure[]
	{\includegraphics[width=0.123\linewidth]{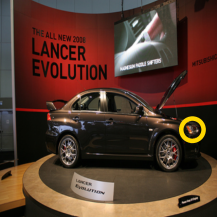}}\hfill
    \subfigure[]
	{\includegraphics[width=0.123\linewidth]{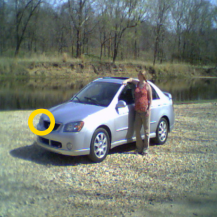}}\hfill
    \subfigure[]
	{\includegraphics[width=0.123\linewidth]{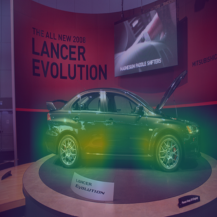}}\hfill
    \subfigure[]
	{\includegraphics[width=0.123\linewidth]{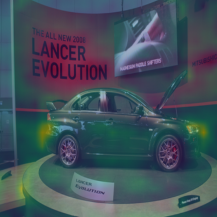}}\hfill
    \subfigure[]
	{\includegraphics[width=0.123\linewidth]{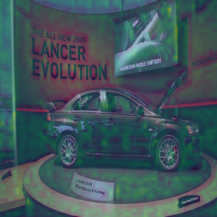}}\hfill
    \subfigure[]
	{\includegraphics[width=0.123\linewidth]{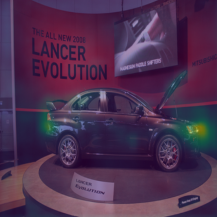}}\hfill
    \subfigure[]
	{\includegraphics[width=0.123\linewidth]{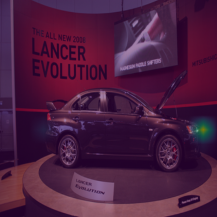}}\hfill
    \subfigure[]
	{\includegraphics[width=0.123\linewidth]{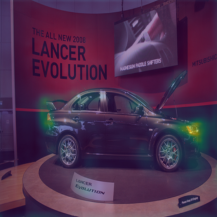}}\hfill\\
	\vspace{-20.5pt}
    \subfigure[]
	{\includegraphics[width=0.123\linewidth]{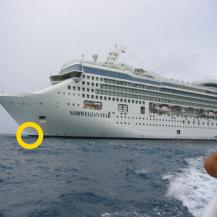}}\hfill
    \subfigure[]
	{\includegraphics[width=0.123\linewidth]{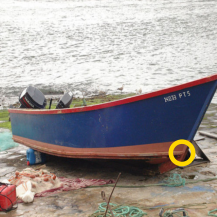}}\hfill
    \subfigure[]
	{\includegraphics[width=0.123\linewidth]{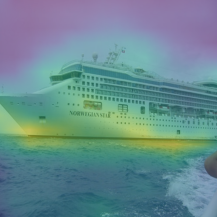}}\hfill
    \subfigure[]
	{\includegraphics[width=0.123\linewidth]{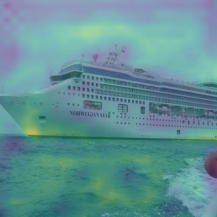}}\hfill
    \subfigure[]
	{\includegraphics[width=0.123\linewidth]{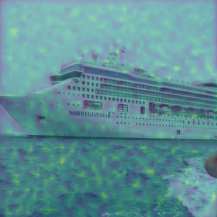}}\hfill
    \subfigure[]
	{\includegraphics[width=0.123\linewidth]{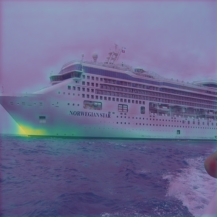}}\hfill
    \subfigure[]
	{\includegraphics[width=0.123\linewidth]{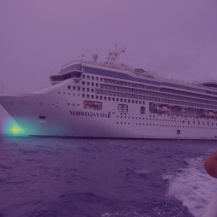}}\hfill
    \subfigure[]
	{\includegraphics[width=0.123\linewidth]{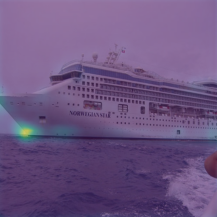}}\hfill\\
	\vspace{-20.5pt}
    \subfigure[]
	{\includegraphics[width=0.123\linewidth]{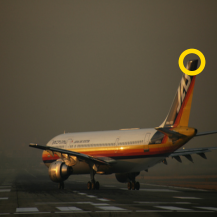}}\hfill
    \subfigure[]
	{\includegraphics[width=0.123\linewidth]{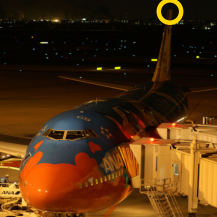}}\hfill
    \subfigure[]
	{\includegraphics[width=0.123\linewidth]{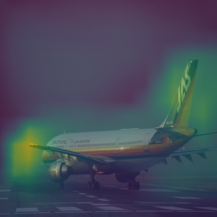}}\hfill
    \subfigure[]
	{\includegraphics[width=0.123\linewidth]{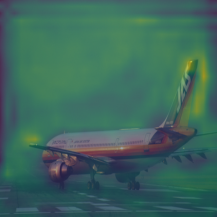}}\hfill
    \subfigure[]
	{\includegraphics[width=0.123\linewidth]{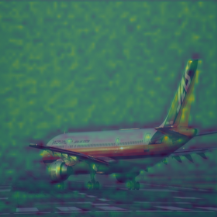}}\hfill
    \subfigure[]
	{\includegraphics[width=0.123\linewidth]{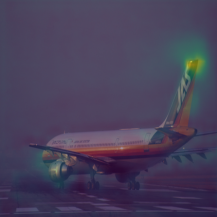}}\hfill
    \subfigure[]
	{\includegraphics[width=0.123\linewidth]{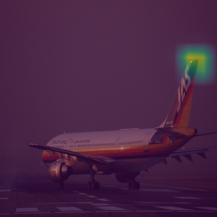}}\hfill
    \subfigure[]
	{\includegraphics[width=0.123\linewidth]{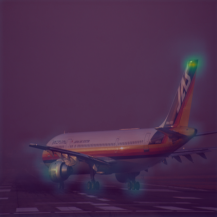}}\hfill\\
	\vspace{-20.5pt}
    \subfigure[]
	{\includegraphics[width=0.123\linewidth]{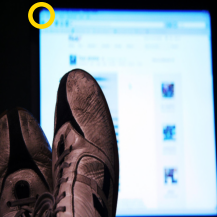}}\hfill
    \subfigure[]
	{\includegraphics[width=0.123\linewidth]{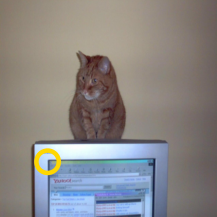}}\hfill
    \subfigure[]
	{\includegraphics[width=0.123\linewidth]{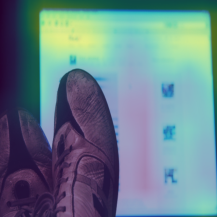}}\hfill
    \subfigure[]
	{\includegraphics[width=0.123\linewidth]{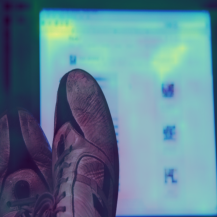}}\hfill
    \subfigure[]
	{\includegraphics[width=0.123\linewidth]{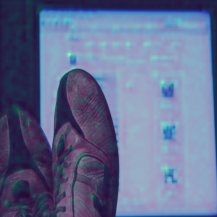}}\hfill
    \subfigure[]
	{\includegraphics[width=0.123\linewidth]{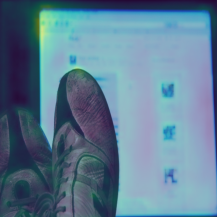}}\hfill
    \subfigure[]
	{\includegraphics[width=0.123\linewidth]{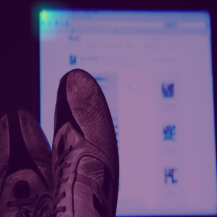}}\hfill
    \subfigure[]
	{\includegraphics[width=0.123\linewidth]{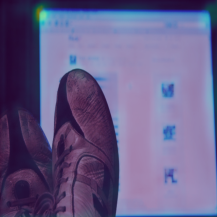}}\hfill\\
	\vspace{-20.5pt}
    \subfigure[]
	{\includegraphics[width=0.123\linewidth]{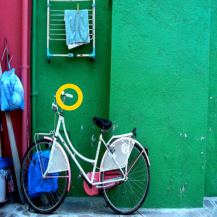}}\hfill
    \subfigure[]
	{\includegraphics[width=0.123\linewidth]{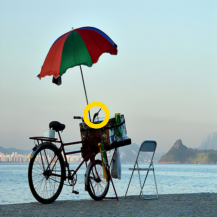}}\hfill
    \subfigure[]
	{\includegraphics[width=0.123\linewidth]{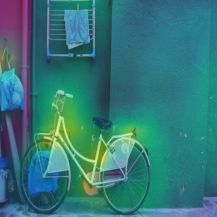}}\hfill
    \subfigure[]
	{\includegraphics[width=0.123\linewidth]{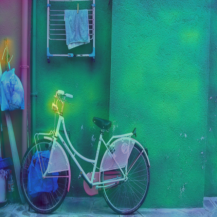}}\hfill
    \subfigure[]
	{\includegraphics[width=0.123\linewidth]{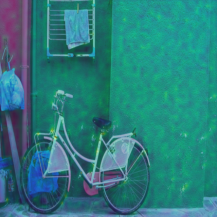}}\hfill
    \subfigure[]
	{\includegraphics[width=0.123\linewidth]{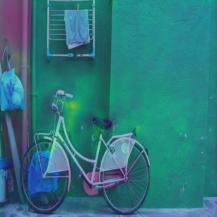}}\hfill
    \subfigure[]
	{\includegraphics[width=0.123\linewidth]{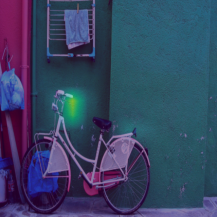}}\hfill
    \subfigure[]
	{\includegraphics[width=0.123\linewidth]{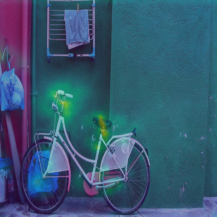}}\hfill\\
	\vspace{-20.5pt}
    \subfigure[]
	{\includegraphics[width=0.123\linewidth]{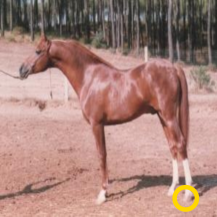}}\hfill
    \subfigure[]
	{\includegraphics[width=0.123\linewidth]{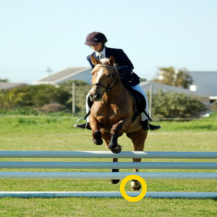}}\hfill
    \subfigure[]
	{\includegraphics[width=0.123\linewidth]{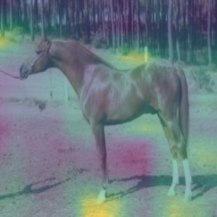}}\hfill
    \subfigure[]
	{\includegraphics[width=0.123\linewidth]{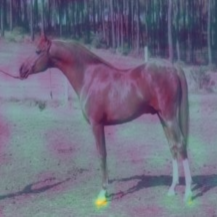}}\hfill
    \subfigure[]
	{\includegraphics[width=0.123\linewidth]{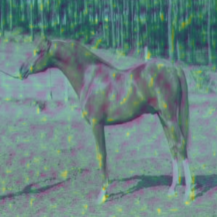}}\hfill
    \subfigure[]
	{\includegraphics[width=0.123\linewidth]{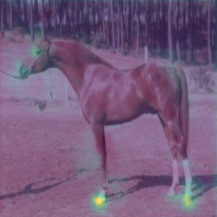}}\hfill
    \subfigure[]
	{\includegraphics[width=0.123\linewidth]{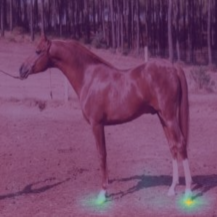}}\hfill
    \subfigure[]
	{\includegraphics[width=0.123\linewidth]{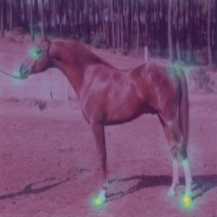}}\hfill\\
	\vspace{-20.5pt}
    \subfigure[(a) Source]
    {\includegraphics[width=0.124\textwidth]{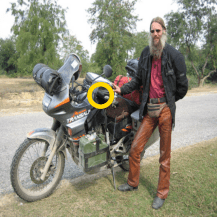}}\hfill
    \subfigure[(b) Target]
    {\includegraphics[width=0.124\textwidth]{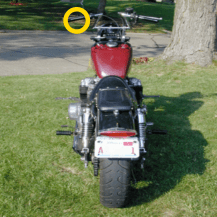}}\hfill
    \subfigure[(c) $C^1$ ]
    {\includegraphics[width=0.124\textwidth]{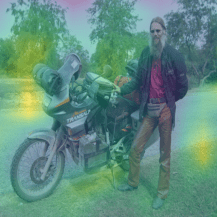}}\hfill
    \subfigure[(d) $C^2$]
    {\includegraphics[width=0.124\textwidth]{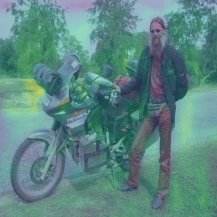}}\hfill
    \subfigure[(e) $C^3$]
    {\includegraphics[width=0.124\textwidth]{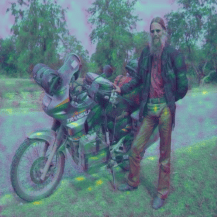}}\hfill
    \subfigure[(f) $D'_s$ $\cdot$ $D'_t$]
    {\includegraphics[width=0.124\textwidth]{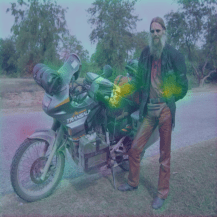}}\hfill
    \subfigure[(g) $C'$]
    {\includegraphics[width=0.124\textwidth]{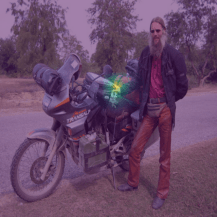}}\hfill
    \subfigure[(h) $D''_s$ $\cdot$ $D''_t$]
    {\includegraphics[width=0.124\textwidth]{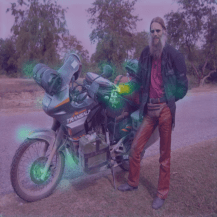}}\hfill\\
	
	\vspace{-5pt}
    \caption{\textbf{Visualization of attention maps.} }\label{attention}\vspace{-10pt}
\end{figure}

\paragraph{More Qualitative Results.}
We provide more qualitative results for SPair-71k~\cite{min2019spair} in Fig.~\ref{fig:spair_quali}, PF-PASCAL~\cite{ham2017proposal} in Fig.~\ref{fig:pascal_quali}, PF-WILLOW~\cite{ham2017proposal} in Fig.~\ref{fig:willow_quali}, and HPatches~\cite{balntas2017hpatches} in Fig.~\ref{hpatches}. We also present more visualization of the attention maps in Fig.~\ref{attention}.

\section*{Limitations}
An apparent limitation of the proposed method is that as IFCAT acts on the cost volume, it is not feasible to perform cost aggregation at higher resolutions. The maximum resolution that would be accessible to users is 64$^4$. This makes the use of standard transformer for attention computation infeasible in terms of memory consumption. Another limitation may be when given a pair of images that are not relevant or show completely different objects, such that there are no correspondences between views, the proposed method lacks an ability to prevent establishing correspondences. 

\section*{Broader Impact}
Our network can be beneficial in a wide range of applications, including simultaneous localization and mapping (SLAM)~\cite{bailey2006simultaneous}, augmented reality (AR)~\cite{peebles2021gan}, object tracking, structure from motion (SfM)~\cite{schonberger2016structure}, optical flow~\cite{fleet2006optical}, and image editing~\cite{barnes2009patchmatch,kim2018recurrent}. As a future work, we could apply the proposed method to different tasks, including feature matching, segmentation and optical flow. Our work would not pose significantly malicious threats on its own. 


\begin{algorithm*}[t]
\caption{Pseudo-Code, PyTorch-like}
\label{1}
\definecolor{codeblue}{rgb}{0.25,0.5,0.5}
\definecolor{codekw}{rgb}{0.85, 0.18, 0.50}

\lstset{
  backgroundcolor=\color{white},
  basicstyle=\fontsize{7.5pt}{7.5pt}\ttfamily\selectfont,
  columns=fullflexible,
  breaklines=true,
  captionpos=b,
  commentstyle=\fontsize{7.5pt}{7.5pt}\color{codeblue},
  keywordstyle=\fontsize{7.5pt}{7.5pt}\color{codekw},
}
\begin{lstlisting}[language=python]

class TransformerLayer:
    def forward(corr, src_feat, trg_feat):
        corr_src, src_feat_refined = integrative_self_attention(corr, src_feat)
        corr_trg, trg_feat_refined = integrative_self_attention(transpose(corr), trg_feat)

        corr = corr_src + transpoze(corr_trg)
        corr = corr + conv4d_1(cost_computation(src_feat_refined, trg_feat_refined))
        corr = corr + conv4d_2(corr)

        src_feat_refined, trg_feat_refined = integrative_cross_attention(corr, src_feat_refined, trg_feat_refined)

        corr = corr + conv4d_3(cost_computation(src_feat_refined, trg_feat_refined))
        corr = corr + conv4d_4(corr)

        return corr, src_feat_refined, trg_feat_refined

class IFCAT:
    def forward(trg_img, src_img):
        src_feat_list = feature_backbone(src_img)
        trg_feat_list = feature_backbone(trg_img)
        src_feats = projection(src_feat_list)
        trg_feats = projection(trg_feat_list)

        correlations = []


        corr_1 = correlation(src_feat_list[0], src_feat_list[0])
        corr_1 = conv4d_1(corr_1)
        src_feat_1, trg_feat_1 = src_feats[0], trg_feats[0]
        corr_1, src_feat_1, trg_feat_1 = transformer_layer[0](corr_1, src_feat_1, trg_feat_1)
        correlations.append(cost_computation(src_feat_1, trg_feat_1))


        corr_2 = correlation(src_feat_list[1], src_feat_list[1]
        corr_2 = corr_1 + conv4d_2(corr_2)

        src_feat_2 = interpolate(linear_1(src_feat_1), scale_factor=2) + src_feats[1]
        trg_feat_2 = interpolate(linear_1(trg_feat_1), scale_factor=2) + trg_feats[1]
        corr_2, src_feat_2, trg_feat_2 = transformer_layer[1](corr_2, src_feat_2, trg_feat_2)
        correlations.append(cost_computation(src_feat_2, trg_feat_2))


        corr_3 = correlation(src_feat_list[2], src_feat_list[2])
        corr_3 = corr_2 + conv4d_3(corr_3)

        src_feat_3 = interpolate(linear_2(src_feat_2), scale_factor=2) + src_feats[2]
        trg_feat_3 = interpolate(linear_2(trg_feat_2), scale_factor=2) + trg_feats[2]
        corr_3, src_feat_3, trg_feat_3 = transformer_layer[2](corr_3, src_feat_3, trg_feat_3)
        correlations.append(cost_computation(src_feat_3, trg_feat_3))


        corr_upsampled = [interpolate4d(x, (64, 64, 64, 64)) for x in correlations]
        c_star = sum(corr_upsampled) / len(corr_upsampled)
        return flow_estimation(c_star)

\end{lstlisting}
\end{algorithm*}

\clearpage

\newpage

\end{document}